\documentclass[letterpaper]{article} 
\usepackage{aaai25}  
\usepackage{times}  
\usepackage{helvet}  
\usepackage{courier}  
\usepackage[hyphens]{url}  
\usepackage{graphicx} 
\usepackage{natbib}  
\usepackage{caption} 
\usepackage{algorithm}
\usepackage{algorithmic}

\usepackage{soul}
\usepackage{url}
\usepackage[utf8]{inputenc}
\usepackage{graphicx}
\usepackage{booktabs}
\usepackage[finalizecache=false,frozencache=true,newfloat]{minted}
\usepackage{mathtools}
\usepackage{amsfonts}
\usepackage{nicefrac}
\usepackage{multirow}
\usepackage{lstautogobble}  
\usepackage{amsmath,amssymb,amsthm}
\usepackage{mdframed}
\usepackage{mathtools}
\usepackage{textcmds}
\usepackage{xspace}
\usepackage{xcolor}
\usepackage{thmtools}
\usepackage{bm}
\usepackage{thm-restate}
\usepackage[inline]{enumitem}
\usepackage{physics}
\usepackage{booktabs} 
\usepackage{tikz}
\usepackage{circuitikz}
\usepackage{dsfont}
\usetikzlibrary{arrows.meta}
\usetikzlibrary{patterns.meta}
\usetikzlibrary{calc,shapes,backgrounds,fit,arrows,positioning,decorations.pathmorphing,patterns}

\usepackage{sourcecodepro}

\definecolor{red_salsa}{HTML}{F94144}
\definecolor{orange_red}{HTML}{F3722C}
\definecolor{yellow_orange}{HTML}{F8961E}
\definecolor{mango_tango}{HTML}{F9844A}
\definecolor{maize_crayola}{HTML}{F9C74F}
\definecolor{pistachio}{HTML}{90BE6D}
\definecolor{jungle_green}{HTML}{43AA8B}
\definecolor{grassy_green}{HTML}{61b876}
\definecolor{steel_teal}{HTML}{4D908E}
\definecolor{queen_blue}{HTML}{577590}
\definecolor{celadon_blue}{HTML}{277DA1}

\definecolor{soft_white}{HTML}{F1F1F1}
\definecolor{softer_white}{HTML}{E5E5E5}
\definecolor{grey}{HTML}{A5A5A5}
\definecolor{dark_grey}{HTML}{8E8E8E}
\definecolor{soft_black}{HTML}{101010}
\definecolor{softer_black}{HTML}{2A2A2A}

\tikzset{
    circnode/.style={
        fill=grey, align=center,
        fill opacity=0.25, text opacity=1,
        circle, 
        minimum width=2.5em
    },
    sqnode/.style={
        fill=grey,
        fill opacity=0.25, text opacity=1,
        rounded corners=0.50em,align=center,
        minimum width=2.5em, minimum height=2.5em
    },
    regpath/.style={
        -Triangle[round], draw=softer_black, ultra thick, rounded corners=0.5em
    },
    thinpath/.style={
        -Triangle[round], draw=softer_black, thick, rounded corners=0.5em
    }
}

\tikzset{
    figpgms/.style={
        inner sep=0mm,
        minimum width=10mm,
        minimum height=10mm
    }
}

\tikzdeclarepattern{
  name=custom_line,
  parameters={
      \pgfkeysvalueof{/pgf/pattern keys/size},
      \pgfkeysvalueof{/pgf/pattern keys/angle},
      \pgfkeysvalueof{/pgf/pattern keys/line width},
  },
  bounding box={
    (0,-0.5*\pgfkeysvalueof{/pgf/pattern keys/line width}) and
    (\pgfkeysvalueof{/pgf/pattern keys/size},
0.5*\pgfkeysvalueof{/pgf/pattern keys/line width})},
  tile size={(\pgfkeysvalueof{/pgf/pattern keys/size},
\pgfkeysvalueof{/pgf/pattern keys/size})},
  tile transformation={rotate=\pgfkeysvalueof{/pgf/pattern keys/angle}},
  defaults={
    size/.initial=5pt,
    angle/.initial=90,
    line width/.initial=7pt,
    xshift/.initial=1.1*\n cm,
    distance/.initial=3pt,
  },
  code={
      \draw [line width=\pgfkeysvalueof{/pgf/pattern keys/line width}]
        (0,0) -- (\pgfkeysvalueof{/pgf/pattern keys/size},0);
  },
}

\theoremstyle{definition}

\theoremstyle{definition}

\theoremstyle{definition}

\theoremstyle{definition}
\newtheorem{example}{Example}[section]
\theoremstyle{definition}

\newcommand{\dspl}{DeepSeaProbLog\xspace}

\newcommand{\mm}{NeSy-MM\xspace}
\newcommand{\mms}{NeSy-MMs\xspace}
\newcommand{\minihack}{MiniHack\xspace}
\newcommand{\nethack}{NetHack\xspace}
\newcommand{\dhmm}{Deep-HMM\xspace}
\newcommand{\dhmms}{Deep-HMMs\xspace}

\newcommand{\nn}{\ensuremath{\boldsymbol{\varphi}}}

\declaretheoremstyle[%
  spaceabove=-6pt,%
  spacebelow=6pt,%
  headfont=\normalfont\itshape,%
  postheadspace=1em,%
  qed=\qedsymbol%
]{mystyle}

\newcommand{\Exp}[2]{\ensuremath{\mathbb{E}_{#1}\left[#2\right]}}
\newcommand{\indicator}{\ensuremath{\mathds{1}}}

\DeclareMathSymbol{\shortminus}{\mathbin}{AMSa}{"39}

\setminted[problog.py:ProbLogLexer -x]{
xleftmargin=0pt,
linenos=false,
escapeinside=@@,
mathescape=true,
breaklines=true,
breakautoindent=True
fontsize=\small
}

\newminted[problog]{problog.py:ProbLogLexer -x}{}
\newcommand{\probloginline}[1]{{\footnotesize\mintinline{problog.py:ProbLogLexer -x}{#1}}}
\newcommand{\mathproblog}[1]{\text{{\footnotesize\mintinline{problog.py:ProbLogLexer -x}{#1}}}}

\usepackage[all, british]{foreign}  
\usepackage{subcaption}             

\title{Relational Neurosymbolic Markov Models}
\author {
    Lennert De Smet\footnote{Equal first authorship}\textsuperscript{\rm 1},
    Gabriele Venturato\footnotemark[1]\textsuperscript{\rm 1},
    Luc De Raedt\footnote{Equal last authorship}\textsuperscript{\rm 1,2},
    Giuseppe Marra\footnotemark[2]\textsuperscript{\rm 1}
}
\affiliations {
    \textsuperscript{\rm 1}KU Leuven, Belgium\\
    \textsuperscript{\rm 2}\"Orebro University, Sweden\\
    firstname.lastname@kuleuven.be
}

\begin{document}

\maketitle

\begin{abstract}
Sequential problems are ubiquitous in AI, such as in reinforcement learning or natural language processing.
State-of-the-art deep sequential models, like transformers, excel in these settings but fail to guarantee the satisfaction of constraints necessary for trustworthy deployment.
In contrast, neurosymbolic AI (NeSy) provides a sound formalism to enforce constraints in deep probabilistic models but scales exponentially on sequential problems.
To overcome these limitations, we introduce relational neurosymbolic Markov models (\mms), a new class of end-to-end differentiable sequential models that integrate and provably satisfy relational logical constraints.
We propose a strategy for inference and learning that scales on sequential settings, and that combines approximate Bayesian inference, automated reasoning, and gradient estimation.
Our experiments show that \mms can solve problems beyond the current state-of-the-art in neurosymbolic AI and still provide strong guarantees with respect to desired properties. Moreover, we show that our models are more interpretable and that constraints can be adapted at test time to out-of-distribution scenarios.
\end{abstract}

\section{Introduction}

Markov models are the theoretical foundation for many successful applications of artificial intelligence,
such as speech recognition~\citep{juang1991hidden}, meteorological predictions~\citep{khiatani2017weather}, games~\citep{schrittwieser2020mastering}, music generation~\citep{austin2021structured}, sports analytics~\citep{van2023markov} and many more~\citep{Mor2020ASR}.
They are so popular mainly because they naturally factorise a sequential problem into step-wise probability distributions.
Such a decomposition leads to better predictions in terms of bias and variance compared to models that do not incorporate the sequential nature of the problem~\citep{bishop2006pattern}.

Neurosymbolic AI (NeSy) has also enjoyed a tremendous increase in attention.
Its general goal is to combine the generalisation potential of symbolic, \ie logical, reasoning with the representational learning prowess of neural networks. This integration leads to interpretable models that can provably satisfy logical constraints.
For example, to guarantee the safety of an autonomous agent~\citep{ijcai2023p637}, to constrain autoregressive language generation~\citep{pmlr-v202-zhang23g} or to impose physical modelling into temporal forecasting~\citep{reichstein2019deep}.

Such a combination already exists in many different flavours, using either fuzzy logic~\citep{badreddine2022logic} or probabilistic logic~\citep{manhaeve2021neural,yang2020neurasp,huang2021scallop,de2023neural},
and either propositional or relational logic~\citep{marra2024statistical}.
The probabilistic case is of special interest, as probabilistic NeSy systems provide a sound semantics to handle uncertainty, as well as to tackle generative tasks.
The relational case is also of special interest as relational logic is a popular and very expressive representation for representing states in, for instance, databases and planning~\citep{RussellNorvig2020}.
Moreover, relational representations facilitate strong generalisation behaviour~\citep{hummel2003symbolic}.
Unfortunately, existing probabilistic or relational NeSy models can not exploit the sequential decomposition inherent to temporal reasoning tasks, thereby limiting their applicability in complex sequential problems.
Therefore, there are still no inference algorithms for such NeSy models that are tailored to scale in sequential settings.

In order to overcome these limitations, we identify four desiderata that a model and its inference algorithm should satisfy.
\begin{enumerate*}[label={\textbf{(D.\Roman*)}}]
    \item It must be able to model and exploit relational logical constraints on states and transition functions.
    It should use relational states as in planning, and ideally it can cope with both continuous and discrete aspects of reality.
    \label{prop:logic}
    \item It has to exploit sequential dependencies without restricting the modelling power, allowing it to scale further than existing NeSy systems.
    \label{prop:sequential}
    \item It must be properly neurosymbolic, that is, it must support transition functions that are logical, neural or purely probabilistic in nature, or any combination thereof.
    Moreover, it must be end-to-end differentiable to allow for the optimisation of any neural components of the model.
    \label{prop:nesy}
    \item It can tackle both discriminative and generative tasks in a probabilistic fashion.
    \label{prop:discgen}
\end{enumerate*}

Both existing probabilistic techniques and neurosymbolic AI are insufficient.
On the neurosymbolic side, scalability~\ref{prop:sequential} remains the biggest problem, and generative capabilities~\ref{prop:discgen} are also lacking.
Purely exact techniques~\citep{manhaeve2021neural,yang2020neurasp} do not scale to non-trivial time horizons, while approximate techniques~\citep{huang2021scallop,van2024nesi} are still limited, \eg they are statistically biased, lack guarantees, and do not support generative tasks.
Only few exact NeSy systems can tackle generative tasks~\citep{de2023neural,misino2022vael} and those that can, are very limited in scalability.
On the side of probabilistic models, only desideratum~\ref{prop:discgen} can be fully met.
Non-parametric techniques can infer any generic hidden Markov model~\citep{koller2009probabilistic} and have been applied in the statistical relational setting~\citep{nitti2016probabilistic}.
However, their integration with the neural paradigm is often paired with strong distributional assumptions~\citep{krishnan2017structured}, such as requiring Gaussian densities.
In particular, gradient-based optimisation is often difficult for general approximate Bayesian inference methods~\citep{scibior2021differentiable,corenflos2021differentiable,younis2023differentiable}.

To fulfil all desiderata, we introduce \emph{relational neurosymbolic Markov models} (\mms), the first integration of deep sequential probabilistic models with NeSy. In particular, 
\begin{enumerate*}[label={(\roman*)}]
    \item we provide a formal definition of \mms,
    \item we introduce a new differentiable neurosymbolic particle filter that combines Rao-Blackwellised~\citep{liu2019rao} inference and state-of-the-art discrete gradient estimation,
    \item we provide an implementation of such models\footnote{The code is available in the supplementary materials, and it will be made publicly available upon acceptance.}, and
    \item we introduce two new benchmarks for generative and discriminative learning, and run an extensive experimental analysis on both.
\end{enumerate*}
The results show that \mms satisfy all the desiderata~\ref{prop:logic} -~\ref{prop:discgen}.

\section{Preliminaries}

\subsection{Markov Models}
Hidden Markov models (HMMs) are sequential probabilistic models for discrete-time Markov processes~\citep{baum1966statistical}. Given sequences of \emph{states} $\mathbf{X} = (\mathbf{X}_t)_{t\in\mathbb{N}}$ and \emph{observations} $\mathbf{Z} = (\mathbf{Z}_t)_{t\in\mathbb{N}}$, an HMM factorises the joint probability distribution $p(\mathbf{X}, \mathbf{Z})$ as,
\begin{align}
\label{eq:hmm}
p(\mathbf{X}_0)p(\mathbf{Z}_0 \mid \mathbf{X}_0)\prod_{t\in\mathbb{N}} p(\mathbf{X}_{t + 1} \mid \mathbf{X}_{t})p(\mathbf{Z}_{t+1} \mid \mathbf{X}_{t+1}),
\end{align}
where $\mathbf{X}_t$ is a fully latent state (Figure~\ref{fig:pgms:hmm}).
If $\mathbf{X}_t$ has a known factorisation in the form of a Bayesian network (BN)~\citep{pearl1988probabilistic}, then the process and its observations encode a \emph{Markovian dynamic Bayesian network} (DBN)~\citep{dean1989model}.
Note that $\mathbf{X}_t$ and $\mathbf{Z}_t$ are random vectors that can have both discrete and continuous components.
In all that follows, a specific assignment of a random variable or vector will be written in lowercase.
For example, $\boldsymbol{x}_t = (x_{t, 1}, \dots, x_{t, D})$ is an assignment of the random vector $\mathbf{X}_t = (X_{t, 1}, \dots, X_{t, D})$ of dimension $D$.

\subsection{Probabilistic Neurosymbolic AI}
            
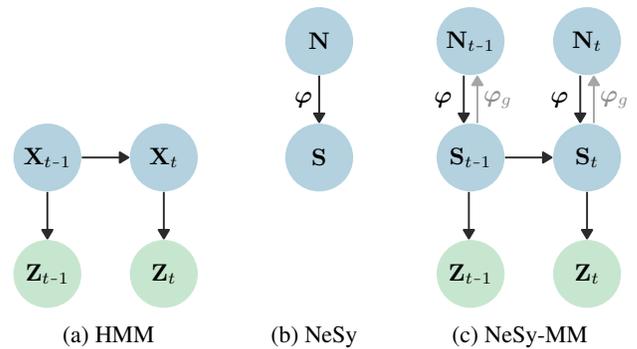
\begin{figure}
\begin{subfigure}{.35\columnwidth}
    \centering
    \resizebox{!}{4cm}{%
    \begin{tikzpicture}[figpgms, node distance=7mm]
        \node[circnode, fill=celadon_blue, fill opacity=0.35] (Xt-1) {$\mathbf{X}_{t\shortminus1}$};
        \node[circnode, fill=grassy_green, fill opacity=0.35, below=of Xt-1] (Zt-1) {$\mathbf{Z}_{t\shortminus1}$};
        \node[circnode, fill=celadon_blue, fill opacity=0.35, right=of Xt-1] (Xt) {$\mathbf{X}_t$};
        \node[circnode, fill=grassy_green, fill opacity=0.35, right=of Zt-1] (Zt) {$\mathbf{Z}_t$};
        \node[above=of Xt-1] {};

        \draw[thinpath] (Xt-1) -- (Zt-1);
        \draw[thinpath] (Xt) -- (Zt);
        \draw[thinpath] (Xt-1) -- (Xt);
    \end{tikzpicture}
    }%
    \caption{HMM}
    \label{fig:pgms:hmm}
\end{subfigure}%
\begin{subfigure}{.3\columnwidth}
    \centering
    \resizebox{!}{4cm}{%
    \begin{tikzpicture}[figpgms, node distance=7mm]
        \node[circnode, fill=celadon_blue, fill opacity=0.35] (N) {$\mathbf{N}$};
        \node[circnode, fill=celadon_blue, fill opacity=0.35, below=of N] (S) {$\mathbf{S}$};
        \node[below=of S] {};

        \draw[thinpath] (N) -- node[midway, left, xshift=0.8em] {$\boldsymbol{\varphi}$} (S);
    \end{tikzpicture}
    }%
    \caption{NeSy}
    \label{fig:pgms:nesy}
\end{subfigure}%
\begin{subfigure}{.35\columnwidth}
    \centering
    \resizebox{!}{4cm}{%
    \begin{tikzpicture}[figpgms, node distance=7mm]
        \node[circnode, fill=celadon_blue, fill opacity=0.35] (Nt-1) {$\mathbf{N}_{t\shortminus1}$};
        \node[circnode, fill=celadon_blue, fill opacity=0.35, below=of Nt-1] (St-1) {$\mathbf{S}_{t\shortminus1}$};   
        \node[circnode, fill=grassy_green, fill opacity=0.35, below=of St-1] (Zt-1) {$\mathbf{Z}_{t\shortminus1}$};
        \node[circnode, fill=celadon_blue, fill opacity=0.35, right=of Nt-1] (Nt) {$\mathbf{N}_t$};
        \node[circnode, fill=celadon_blue, fill opacity=0.35, right=of St-1] (St) {$\mathbf{S}_t$};
        \node[circnode, fill=grassy_green, fill opacity=0.35, right=of Zt-1] (Zt) {$\mathbf{Z}_t$};

        \draw[thinpath] ([xshift=-1mm]Nt-1.south) -- node[midway, left, xshift=0.6em] {$\boldsymbol{\varphi}$} ([xshift=-1mm]St-1.north);
        \draw[thinpath, grey] ([xshift=1mm]St-1.north) -- node[midway, right, xshift=-0.6em] {\textcolor{dark_grey}{$\boldsymbol{\varphi}_{g}$}} ([xshift=1mm]Nt-1.south);

        \draw[thinpath] ([xshift=-1mm]Nt.south) -- node[midway, left, xshift=0.6em] {$\boldsymbol{\varphi}$} ([xshift=-1mm]St.north);
        \draw[thinpath, grey] ([xshift=1mm]St.north) -- node[midway, right, xshift=-0.6em] {\textcolor{dark_grey}{$\boldsymbol{\varphi}_{g}$}} ([xshift=1mm]Nt.south);
        
        \draw[thinpath] ([xshift=-0.25mm]St-1.south) -- ([xshift=-0.25mm]Zt-1.north);
        
        \draw[thinpath] (St) -- (Zt);

        \draw[thinpath] (St-1) -- (St);
    \end{tikzpicture}
    }%
    \caption{\mm}
    \label{fig:pgms:nesy-mm}
\end{subfigure}%
\caption{Probabilistic graphical model representations of the different systems considered in this work. Blue represents the states, green the observations.
}
\label{fig:pgms}
\end{figure}

Probabilistic NeSy methods originate from the field of statistical relational AI (StarAI) that integrates statistical AI with logic~\citep{raedt2016statistical, marra2024statistical}. This integration leads to systems capable of performing inference and learning with uncertainty over \emph{symbolic}, \ie logical, knowledge.
For example, the logical relations \probloginline{player_at(player1, location1)} and \probloginline{monster_at(monster1, location1)} can be used to apply the rule \probloginline{hit(P,M) :- player_at(P,L), monster_at(M,L)} and deduce that \probloginline{player1} is hit by \probloginline{monster1} because they are in the same location.
Moreover, this knowledge is often uncertain in practice, resulting in uncertainty on whether the deduced logical relations hold.
For example, consider the case of a sneaky monster.
If we are unsure whether \probloginline{monster_at(monster1, location1)} is true or not, we will also be uncertain whether the player is hit or not.
Notice that uncertain logical relations can be modelled as binary random variables, which justifies the integration with statistics.

While StarAI assumes knowledge to be neatly represented as a symbolic state $\mathbf{S}$, such an assumption does not always hold. 
Images, sound waves or natural language are usually represented as \emph{subsymbolic} data, \ie tensors, that are not directly usable by relational AI.
Therefore, probabilistic NeSy methods use neural predicates $\boldsymbol{\varphi}$ to map subsymbolic data to a probability distribution over symbolic representations that can be used by StarAI.
Figure~\ref{fig:pgms:nesy} depicts a probabilistic graphical model (PGM)~\citep{koller2009probabilistic} representation of a NeSy system.
More formally, given a boolean variable $Y$ from $\mathbf{S}$ with domain $\{y, \neg y\}$ and a set of rules $\mathcal{R}$ on the symbols in $\mathbf{S}$, inference in NeSy computes the probability that the query $Y$ is true via weighted model integration (WMI)~\citep{morettin2021hybrid}
\begin{align}\label{eq:nesy-inference}
p_{\boldsymbol{\varphi}}(y \mid \mathbf{n}) 
&=\int \indicator_{\mathbf{s} \models_{\mathcal{R}} y}\ p_{\boldsymbol{\varphi}}(\mathbf{s} \mid \mathbf{n})
\ \mathrm{d}\mathbf{s},
\end{align}
where the distribution of $\mathbf{S}$ is parametrised by a neural network $\boldsymbol{\varphi}$ from the subsymbolic state $\mathbf{n}$.

A prominent way of representing neurosymbolic models is via probabilistic logic programming (PLP)~\citep{de2007problog}.
A running example will illustrate the main concepts, while a more technical exposition is given in Appendix~\ref{app:plp}.

\begin{figure}
{\fontsize{8.5}{9.5}\selectfont
\begin{problog}
player(Im, P) ~ normal(noisy_player(Im)).
monster(Im, M) ~ normal(noisy_monster(Im)).
clumsy ~ bernoulli(0.75).

hits(M, P) :-
   distance(M, P, D), D < 2, not clumsy.
game_over(Im) :-
   player(Im, P), monster(Im, M), hits(M, P).
\end{problog}
}
\caption{\dspl~\citep{de2023neural} encoding of Example~\ref{ex:nesy-game}.
The first two lines are \emph{neural predicates} that represent deep random variables modelling the normally distributed locations of the monster \probloginline{M} and the player \probloginline{P}.
Each of the neural predicates has a named neural network that takes the image \probloginline{Im} as input and outputs the parameters of its random variable.
The third line introduces a Bernoulli random variable \probloginline{clumsy} indicating that the monster will be clumsy with a $75\%$ chance.
The final two lines express two rules of the game that determine when the monster \probloginline{M} \probloginline{hits} the player \probloginline{P} and when the image \probloginline{Im} depicts a lost game, \ie when the image depicts the monster hitting the player.
}
\label{fig:nesy-game}
\end{figure}

\begin{example}
\label{ex:nesy-game}
Consider a simple game where a monster \probloginline{M} and player \probloginline{P} interact with each other (Figure~\ref{fig:nesy-game}).
Both entities are represented by their normally distributed locations, which are parametrised by neural networks from a given image \probloginline{Im}.
Their interactions are in the form of \probloginline{hits}, where the monster can hit the player if it is close and not \probloginline{clumsy}.
The clumsiness of the monster is uncertain and modelled by a separate Bernoulli random variable.
Finally, the game ends whenever the monster succeeds in hitting the player. 
Because of the uncertainty on locations and clumsiness, it also follows that whether the game is over is uncertain.
\end{example}

\section{Relational Neurosymbolic Markov Models}
\label{sec:nesy-mm}

\emph{Relational neurosymbolic Markov models (\mms)} combine the sequential and partially observable nature of HMMs and DBNs (Figure~\ref{fig:pgms:hmm}) with neurally parametrised relational probability distributions (Figure~\ref{fig:pgms:nesy}).
That is, we consider Markov processes $\mathbf{X} = (\mathbf{X}_t)_{t \in \mathbb{N}}$ with observations $\mathbf{Z} = (\mathbf{Z}_t)_{t \in \mathbb{N}}$ where the state $\mathbf{X}_t$ is now a \emph{neurosymbolic state} $\mathbf{X}_t = (\mathbf{N}_t, \mathbf{S}_t)$.
Figure~\ref{fig:pgms:nesy-mm} depicts the graphical model of this novel integration.
\mms represent joint probability distributions $p_{\boldsymbol{\varphi}}(\mathbf{N}, \mathbf{S}, \mathbf{Z})$ that factorise as
\begin{align}
    \label{eq:nesy-mm}
    &p_{\boldsymbol{\varphi}}(\mathbf{S}_0 \mid \mathbf{N}_0) p(\mathbf{N}_0) p(\mathbf{Z}_0 \mid \mathbf{S}_0) \notag\\
    &\prod_{t\in\mathbb{N}}
    p_{\boldsymbol{\varphi}}(\mathbf{S_{t + 1}} \mid \mathbf{S}_t, \mathbf{N}_{t + 1}) p(\mathbf{N}_{t+1}) p(\mathbf{Z}_{t+1} \mid \mathbf{S}_{t+1}).
\end{align}

Despite the similarity with Eq.~\ref{eq:hmm}, \mms are complex models that define a wide variety of distributions, taking into account our four desiderate of interest~\ref{prop:logic} - \ref{prop:discgen}.

\paragraph{\mms explicitly model symbols and their relations.}
Having a NeSy state means we perform inference in a symbolic state space where \emph{relational logic rules} $\mathcal{R}$ govern the relationship between symbols, both within a single time slice and in the transition between states.
This relational symbolic space allows \mms to incorporate human knowledge into our reasoning process, giving guarantees on how the sequential process evolves, \eg we can guarantee safety properties throughout the entire sequence (see Example~\ref{ex:markov-game}).
Additionally, the relational aspect significantly enhanced the out-of-distribution generalisation potential (Section~\ref{sec:experiments}).
\paragraph{\mms factorise symbols over sequences.}
Standard NeSy systems (Figure~\ref{fig:pgms:nesy}) must model the full joint distribution over time, \ie $p(\mathbf{S}) = p(\mathbf{S}_1, \ldots, \mathbf{S}_t)$. In contrast, we can factorise the distribution thanks to the Markovian neurosymbolic transition function $p_{\boldsymbol{\varphi}}(\mathbf{S_{t + 1}} \mid \mathbf{S}_t, \mathbf{N}_{t + 1})$, allowing for the definition of probabilistic temporal relations between symbols.
Moreover, such a factorisation dramatically simplifies the symbolic space by exploiting the sequential dependencies that standard NeSy systems ignore.
\paragraph{\mms integrate neural and logical parametrisations.}
The symbols of a \mm and their transitions need not be purely logical and can be parametrised by neural networks.
This flexibility in parametrisation not only bridges the gap between subsymbols and symbols, but also allows for neural networks to fill in gaps in background knowledge.
For example, when faced with learning the behaviour of another entity in a game while being constrained by the rules of the game (Section~\ref{subsec:experiments:transitions}).
In essence, \mms place symbols and logic where knowledge is available, while using neural nets to parametrise symbols and structure where necessary.

\paragraph{\mms express discriminative and generative neurosymbolic models.}
When given a target variable $Y$, which can be any of the symbols in $\mathbf{S}$ or a logical derivation thereof, a \mm can answer conditional \emph{discriminative NeSy queries} of the form $p_{\boldsymbol{\varphi}}(y \mid \mathbf{n}, \mathbf{z})$ via
\begin{align}
    \int\limits_{\mathrlap{\mathbf{s} \models_{\mathcal{R}} y}}
    p_{\boldsymbol{\varphi}}(\mathbf{s}_0 \mid \mathbf{n_0}, \mathbf{z_0})
    \prod_{t\in\mathbb{N}}
    p_{\boldsymbol{\varphi}}(\mathbf{s_{t + 1}} \mid \mathbf{s}_t, \mathbf{n}_{t+1}, \mathbf{z}_{t+1})
    \ \mathrm{d}\mathbf{s}.
\end{align}

Alternatively, we can assume a generative perspective by defining the neural parametrisation of our model with a generative model~\citep{goodfellow2020generative, dinh2016density, ho2020denoising}, \ie the inverted $\boldsymbol{\varphi}_g$ edges in Figure~\ref{fig:pgms:nesy-mm}.
This perspective leads to the factorisation
\begin{align} 
    \int p_{\boldsymbol{\varphi}}(\mathbf{s}_0, \mathbf{N}_0 \mid \mathbf{z}_0)
    \prod_{t\in\mathbb{N}}
    p_{\boldsymbol{\varphi}}(\mathbf{s_{t + 1}}, \mathbf{N}_{t + 1} \mid \mathbf{s}_t, \mathbf{z}_{t+1})
     \ \mathrm{d}\mathbf{s},
\end{align}
of $p_{\boldsymbol{\varphi}}(\mathbf{N} \mid \mathbf{z})$.
That is, \mms can tackle \emph{generative tasks} where samples $\mathbf{n}$ from $p_{\boldsymbol{\varphi}}(\mathbf{N} \mid \mathbf{z})$ that satisfy the possibly logical evidence $\mathbf{z}$ are asked.
We showcase this functionality in Section~\ref{subsec:experiments:results}, where we use a VAE~\citep{kingma2013auto} to generate sequences of images of a game that adhere to the rules of the game.
\begin{figure*}
\begin{minipage}{0.7\linewidth}
{\fontsize{8.5}{9.5}\selectfont
\begin{problog}
player(Im, P)@$_{\texttt{0}}$@ ~ normal(noisy_player(Im)).
player(Im, P)@$_{\texttt{t}}$@ ~ normal(Next) :-
    player(Im, P)@$_{\texttt{t - 1}}$@, monster(Im, M), player_move(P, M, Next).
monster(Im, M) ~ normal(noisy_monster(Im)).
clumsy ~ bernoulli(0.75).

hits(M, P)@$_{\texttt{t}}$@ :- distance(M, P, D)@$_{\texttt{t}}$@, D < 2, not clumsy.

game_over@$_{\texttt{t}}$@(Im) :- player(Im, P)@$_{\texttt{t}}$@, monster(Im, M), hits(M, P)@$_{\texttt{t}}$@.

safe@$_{\texttt{t}}$@(Im, P) :- 
    player(Im, P)@$_{\texttt{t}}$@, monster(Im, M),  distance(M, P, D)@$_{\texttt{t}}$@, D > 2.

observe(safe@$_{\texttt{0:T}}$@, true).
\end{problog}
}
\end{minipage}%
\begin{minipage}{0.30\linewidth}
\centering
\resizebox{0.72\linewidth}{!}{%
\begin{tikzpicture}
\node[sqnode, fill=celadon_blue, fill opacity=0.35] (image) {\texttt{Im}};
\node[circnode, fill=celadon_blue, fill opacity=0.35, below=3em of image] (player) {$\texttt{P}_\texttt{t}$};
\node[circnode, fill=celadon_blue, fill opacity=0.35, right=2em of image] (monster) {\texttt{M}};
\node[circnode, fill=celadon_blue, fill opacity=0.35, below right=0em and 1.25em of monster] (clumsy) {\texttt{C}};
\node[circnode, fill=celadon_blue, fill opacity=0.35, below=3em of monster] (hits) {$\texttt{H}_\texttt{t}$};
\node[circnode, fill=grassy_green, fill opacity=0.35, below=3em of player] (safe) {$\texttt{S}_\texttt{t}$};
\node[circnode, fill=celadon_blue, fill opacity=0.35, right=6.5em of safe] (game_over) {\texttt{G}};

\draw[thinpath] ([xshift=-1mm]image.south) -- node[midway, left, yshift=0.5em] {$\varphi^\texttt{p}$} ([xshift=-1mm]player.north);
\draw[thinpath] (image) -- node[midway, above] {$\varphi^\texttt{m}$} (monster);
\draw[thinpath] (player) -- (hits);
\draw[thinpath] ([xshift=-1mm]player.south) -- ([xshift=-1mm]safe.north);
\draw[thinpath] (monster.south) 
-- ([yshift=-1.40em] monster.south)
-| ([xshift=1em]safe.east)
-- (safe.east);
\draw[thinpath] ([xshift=-0.5mm]monster.south) -- ([xshift=-0.5mm, yshift=-1.25em] monster.south) -|([xshift=1mm]player.north);
\draw[thinpath] ([xshift=0.5mm]monster.south) -- ([xshift=0.5mm]hits.north);
\draw[thinpath] (clumsy) |- (hits);
\draw[thinpath] (hits) |- (game_over.west);
\draw[thinpath] ([xshift=1mm]player.south) -- ([xshift=1mm, yshift=-1.5em] player.south) -| ([xshift=-1mm]game_over.north);
\draw[thinpath] (monster.east) -| ([xshift=1mm]game_over.north);

\begin{pgfonlayer}{background}
        \node[rounded corners=0.50em, fit=(player) (hits) (safe), inner sep=1em, draw=grey, line width=2pt, opacity=0.5] (rect) {};
\end{pgfonlayer}
\node[xshift=-1.5em, yshift=1em, inner sep=2pt] at (rect.south east) {\textcolor{dark_grey}{$\texttt{0:T}$}};
\end{tikzpicture}
}
\end{minipage}
\caption{On the left, a logic programming description of the game Example~\ref{ex:markov-game} in the discrete-continuous probabilistic NeSy language \dspl.
On the right, the corresponding graphical model. We use plate notation to indicate a Markov transition. A rolled-out version is available in the appendix in Figure~\ref{fig:markov-game:rolled-out}.}
\label{fig:markov-game}
\end{figure*}
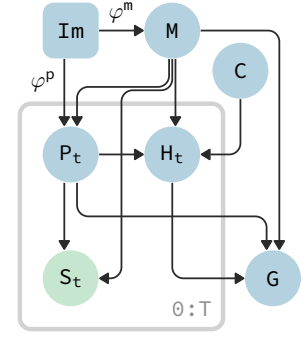

\begin{example}
\label{ex:markov-game}
Figure~\ref{fig:markov-game} shows a new version of the game from Example~\ref{ex:nesy-game}. The player can now move in the environment with a Markovian transition function \probloginline{player_move} based on the player's previous location and the static monster's position.
The observation rule \probloginline{safe} guarantees the player's safety at every time step within the horizon $\texttt{0:T}$.
Finally, we can query \mathproblog{game_lost(image.png)}$_\texttt{t}$ for any $\texttt{t} \in \{\texttt{0,}\ldots\texttt{,T}\}$.
Notice that this \mm depends only on the first image at time $\texttt{t{=}0}$ and that the projection into the future is done via the logical transition rules.
\end{example}

\section{Inference and Learning}
\label{sec:method}
To bridge the gap between NeSy and sequential probabilistic models, we propose a new, differentiable inference technique that combines non-parametric approximate Bayesian inference with exact NeSy inference.
In the following sections, we will distinguish between random variables with finite and infinite domains.
The latter includes both countably infinite and continuous (uncountable) domains.

\subsection{Differentiable \mm Particle Filtering}
Traditional particle filters are not differentiable because they perform resampling (Appendix~\ref{app:nesy-pf}).
Resampling is needed because the observations $\mathbf{Z}_t$ are separated from the transitions $p_{\nn}(\mathbf{X}_{t + 1} \mid \mathbf{X}_{t})$, which means the conditional distribution $p_{\nn}(\mathbf{X}_{t + 1} \mid \mathbf{X}_{t}, \mathbf{Z}_{t + 1})$ is not readily available.
The current state-of-the-art solution is to recover the differentiability of resampling~\citep{scibior2021differentiable,corenflos2021differentiable,younis2023differentiable}.
On the contrary, we propose a novel solution that takes advantage of the neurosymbolic nature of a \mm.
In particular, we circumvent the problem of differentiating through resampling by using a Rao-Blackwellised particle filter (RBPF)~\citep{murphy2001rao}. 
A RBPF assumes $p_{\nn}(\mathbf{X}_{t + 1} \mid \mathbf{X}_{t}, \mathbf{Z}_{t + 1})$ can be computed exactly and uses it to recursively compute $p_{\nn}(\mathbf{X}_{t + 1} \mid \mathbf{Z}_{0:t + 1})$ as
\begin{align}\label{eq:rbpf_recursion}
    \int
    p_{\nn}(\mathbf{X}_{t + 1} \mid \mathbf{x}_{t}, \mathbf{Z}_{t + 1}) p_{\nn}(\mathbf{x}_{t} \mid \mathbf{Z}_{0:t})
    \ \mathrm{d}\mathbf{x}_{t}.
\end{align}
We claim it is viable to compute $p_{\nn}(\mathbf{X}_{t + 1} \mid \mathbf{X}_{t}, \mathbf{Z}_{t + 1})$ in our NeSy setting because, when $\mathbf{X}_t$ is purely discrete, computing these probabilities can leverage the advances in exact inference from both neurosymbolic AI~\citep{kisa2014probabilistic,tsamoura2021materializing} and probabilistic AI~\citep{darwiche2020advance,holtzen2020scaling}.

By removing resampling and having access to the exact transition probabilities, we can exploit an up-until-now unexplored synergy with gradient estimation.
State-of-the-art unbiased discrete gradient estimation algorithms~\citep{kool2019buy,de2023differentiable} use samples and the gradients of the probability of those samples to approximate the gradients of finite distributions.
In other words, they need the exact probabilities of these distributions to function.
Hence, since our RBPF computes $p_{\nn}(\mathbf{X}_{t + 1} \mid \mathbf{X}_{t}, \mathbf{Z}_{t + 1})$ exactly, gradient estimation can be used to recursively get approximate gradients for the distribution $p_{\nn}(\mathbf{X}_{t + 1} \mid \mathbf{Z}_{0:t + 1})$.
For example, using the Log-Derivative trick~\citep{williams1992simple}
\begin{align}\label{eq:log_trick_npf}
    &\nabla_{\nn} p_{\nn}(\mathbf{X}_{t + 1} \mid \mathbf{Z}_{0:t + 1}) \\
    \nonumber
    &=
    \Exp{
        \mathbf{X}_t
    }{
        \nabla_{\nn}
        p_{\nn}(\mathbf{X}_{t + 1} \mid \mathbf{X}_{t}, \mathbf{Z}_{t + 1})
    } \\
    &\hspace{1.1pt}+
    \Exp{
        \mathbf{X}_t
    }{
        p_{\nn}(\mathbf{X}_{t + 1} \mid \mathbf{X}_{t}, \mathbf{Z}_{t + 1})
        \nabla_{\nn}
        \log p_{\nn}(\mathbf{X}_{t} \mid \mathbf{Z}_{0:t})
    } \nonumber
    .
\end{align}
In our implementation, we opted for the state-of-the-art performance of RLOO~\citep{kool2019buy} for gradient estimation.
More precise details on our application of gradient estimation can be found in Appendix~\ref{app:gradients}.

\subsection{NeSy inference via cluster factorisation}
Unfortunately, computing $p_{\nn}(\mathbf{X}_{t + 1} \mid \mathbf{x}_{t}, \mathbf{Z}_{t + 1})$ exactly when $\mathbf{X}_{t + 1}$ also contains variables with an infinite domain is generally only possible under strict assumptions such as Gaussian densities.
Moreover, it can still become prohibitively expensive in the purely finite case when ignoring the internal dependency structure of $\mathbf{X}_{t + 1}$.
We mitigate these problems by factorising the \mm further into clusters of variables that become independent when conditioning on $\mathbf{Z}$.
Specifically, a conditional probability distribution $p(\mathbf{X} \mid \mathbf{Z})$ can be factorised as
\begin{align}
    p(\mathbf{X} \mid \mathbf{Z}) = \prod\limits_{i=1}^B p(\mathbf{X}^i \mid \mathbf{Z}),
\end{align}
where $B$ is the maximal number of clusters. 
Intuitively, variables within the same cluster must always be sampled together and hence comprise minimal subproblems to be solved.
The distribution $p(\mathbf{X}^i \mid \mathbf{Z})$ of each of the subproblems can be computed separately to alleviate the computational bottleneck of computing $p(\mathbf{X} \mid \mathbf{Z})$ exactly.

Applying the cluster factorisation to the conditional probability distribution $p_{\nn}(\mathbf{X}_{t + 1} \mid \mathbf{x}_{t}, \mathbf{Z}_{t + 1})$ with clusters $\{\mathbf{X}_{t + 1}^i\}_{i = 1}^B$ yields
\begin{align}
    p_{\nn}(\mathbf{X}_{t + 1} \mid \mathbf{x}_{t}, \mathbf{Z}_{t + 1})
    =
    \prod_{i = 1}^{B}
    p_{\nn}(\mathbf{X}_{t + 1}^{i} \mid \mathbf{x}_{t}, \mathbf{Z}_{t + 1}).
\end{align}
If we split every cluster $\mathbf{X}_t^i$ into a finite part $\mathbf{F}_t^i$ and infinite part $\mathbf{I}_t^i$, this factorisation can be further refined into
\begin{align}
    \prod_{i = 1}^{B}
    p_{\nn}(\mathbf{F}_{t + 1}^{i} \mid \mathbf{I}_{t + 1}^i \mathbf{x}_{t}, \mathbf{Z}_{t + 1})
    p_{\nn}(\mathbf{I}_{t + 1}^{i} \mid \mathbf{x}_{t}, \mathbf{Z}_{t + 1}).
\end{align}
By first obtaining samples for the infinite random variables $\mathbf{I}_t^{i}$ in every $i^{\text{th}}$ cluster using a traditional particle filter, we are again left with a purely finite probability distribution $p_{\nn}(\mathbf{F}_{t + 1}^{i} \mid \mathbf{I}_{t + 1}^i \mathbf{x}_{t}, \mathbf{Z}_{t + 1})$ that we compute exactly such that discrete gradient estimation can be applied.

For infinite random variables, we can recover differentiability using any of the proven and tailored gradient estimation algorithms~\citep{scibior2021differentiable,corenflos2021differentiable,younis2023differentiable}.
In our case, we followed the work of~\citet{scibior2021differentiable} as it provides strong baseline performance.
In summary, we recover gradient-based optimisation of infinite, finite and binary logical variables by joining local exact inference with specialised gradient estimation.
\emph{The result is a novel Rao-Blackwellised particle filter for \mms that handles hybrid domains and exploits the inner conditional dependency structure of the NeSy states $\mathbf{X}_t$.}

\section{Experiments}
\label{sec:experiments}

In the following two sections, we present our generative and discriminative benchmarks, and show that \mms are capable of tackling both settings~\ref{prop:discgen}.
We will also clearly show how the presence of relational logic in \mms significantly and positively impacts both in- and out-of-distribution performance compared to state-of-the-art deep (probabilistic) models~\ref{prop:logic}.
In doing so, we show that \mms scale to problem settings far beyond the horizon of existing NeSy methods~\ref{prop:sequential}. 
In total, \mms are successful neurosymbolic models capable of optimising various neural components while adhering to logical constraints~\ref{prop:nesy}.

\begin{figure*}[t]
    \centering
    \begin{subfigure}[b]{0.49\textwidth}
        \centering
        \includegraphics[width=\textwidth]{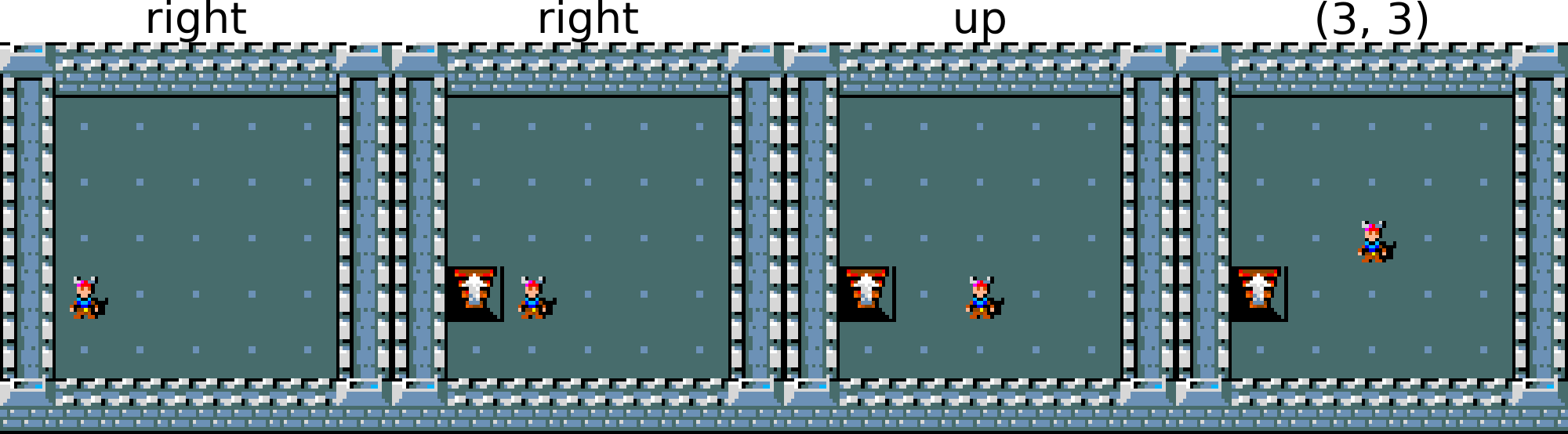}
        \caption{}
        \label{fig:minihack-dataset}
    \end{subfigure}
    \hfill
    \begin{subfigure}[b]{0.49\textwidth}
        \centering
        \includegraphics[width=\textwidth]{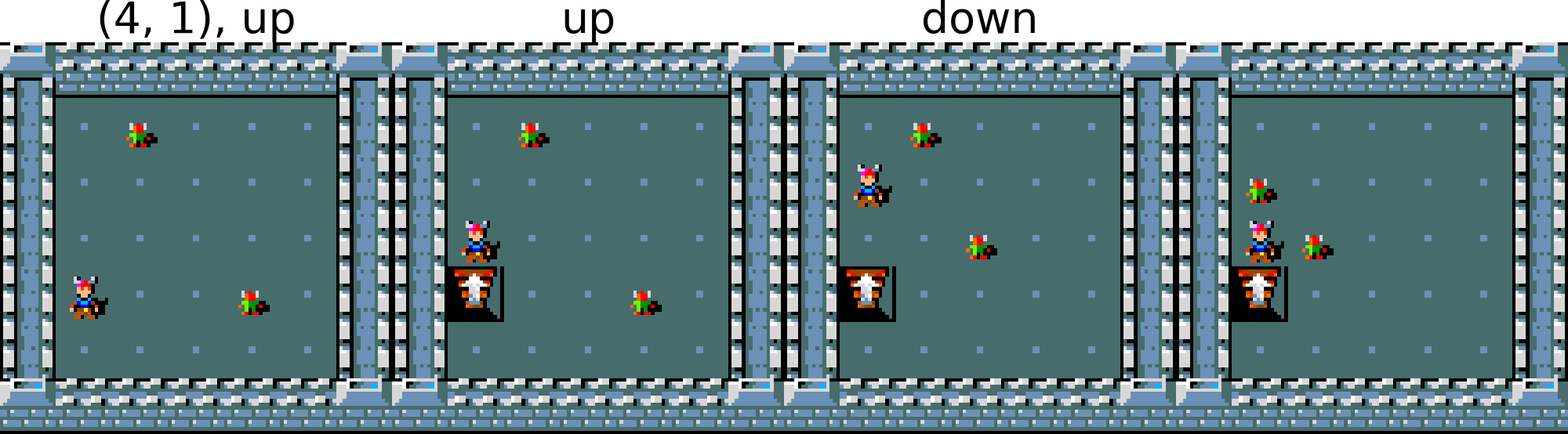}
        \caption{}
        \label{fig:enemyroom-dataset}
    \end{subfigure}
    \caption{
        Example trajectories of length $4$ in a $5{\times}5$ grid for the generative (a) and discriminative (b) datasets, with the corresponding labels above the images.
        Note that for the discriminative task, the models do not take images as input but rather the symbolic state. The images are provided for visualization purposes.
    }
    \label{fig:datasets}
\end{figure*}

\subsection{Generative}
\label{subsec:experiments:generative}

Our generative experiment is inspired by the Mario experiment of \citet{misino2022vael}, extended using MiniHack~\citep{samvelyan2021minihack}, a flexible framework to define environments of the open-ended game NetHack~\citep{kuttler2020nethack}.
The dataset consists of trajectories of images of length $T$ representing an agent moving $T$ steps in a grid world of size $N{\times}N$ surrounded by walls.
The starting position of the agent is randomly initialised and the actions the agent takes are uniformly sampled among the four cardinal directions, \ie up, down, left, right (Figure~\ref{fig:minihack-dataset}).
The actions the agent took at every time step are also given in the trajectory.
During training, the model takes sequences of both \minihack images and actions and learns to reconstruct the given images.
At test time, the model should then be able to generate sequences of \minihack images that follow a given sequence of actions and satisfy the rules of \nethack.

We use VAEL~\citep{misino2022vael} as a neurosymbolic baseline and two other fully neural baselines:
a variational transformer architecture (VT); and a \mm without logical rules and with neural networks as transition function (\dhmm).
Since \dhmms are subsumed by \mms, the baseline was implemented in our framework to allow it to benefit from our Rao-Blackwellised inference and learning strategy.

We consider two metrics for the evaluation. First, the reconstruction error (RE), measured by the mean absolute difference in pixel values, which is first averaged over the images, then separately averaged over all images of the sequence. Second, the reconstruction accuracy (RA), which uses a pre-trained classifier for the location of the agent and measures how much the reconstructed trajectory aligns with the ground truth.
This is crucial to understand whether the agent is moving according to the actual rules of the game.

\subsection{Discriminative}
\label{subsec:experiments:transitions}

The next setting consists of a discriminative task, where the goal is to classify trajectories of symbolic states.
The main challenge is that the transition function is now partially unknown and needs to be learned from examples.
That is, we do not use neural networks for perception as is usually done in NeSy, but have a transition that is both neural and logical.
More concretely, the dataset for the discriminative task consists of trajectories similar to the generative dataset. However, there are now also enemies present that are trying to kill the agent (Figure~\ref{fig:enemyroom-dataset}).
The input to the model in this case is fully symbolic, meaning we do not input images, but rather the precise starting coordinate of the agent and the list of actions performed by the agent.
On top of that, we observe if one of the enemies hits the agent, \ie the observations $Z_t$ are binary random variables.
The discriminative task is binary classification, where a trajectory has label 1 if the agent dies somewhere in the trajectory and 0 otherwise.
While we know the basic rules of \nethack, such as permitted movements and damage mechanics, we do not know the transition function of the enemy. 
That is, we don't know the behaviour of the enemies and fill this gap in knowledge with a neural network that should respect the known rules of \nethack (see Appendix~\ref{app:reproducibility}).
We assume that all the enemies share the same behaviour.

Similar to the generative experiment, we use a transformer and a \dhmm as baselines, this time in a discriminative configuration (Appendix~\ref{app:reproducibility}).
To specifically gauge the out-of-distribution (OOD) generalisation capabilities of all methods, we train only using simple sequences of length 10 containing just one enemy moving on a $10{\times}10$ grid and we test on more complex sequences.
The OOD cases consider different combinations of sequences on grids of size $10{\times}10$ or $15{\times}15$, length 10 or 20, and with 1 or 2 enemies.
When more enemies are present or the sequences are longer, it is naturally easier for the agent to be killed.
Conversely, the enemies might need more steps to reach the agent when the grid is bigger.
These differences lead to a difference in class balance from one configuration to the other (Table~\ref{tab:death_percentage}), posing an additional significant learning challenge.
Ideally, we want a model that is able to counterbalance the bias inherent to its training data.
To test such abilities, we evaluate all methods in terms of both balanced accuracy and F1-score.

\begin{table}[t]
    \renewcommand{\arraystretch}{0.8}
    \centering
    \begin{tabular}{ccccc}
        $N$ & $T$ & $E$ & \textbf{Death (\%)} & \textbf{OOD} \\
        \midrule
        \multirow{4.5}{*}{10} & \multirow{2}{*}{10} & 1 & 17.2 & -- \\
                             &    & 2 & 60.9 & \checkmark \\
        \cmidrule{2-5}
                             & \multirow{2}{*}{20} & 1 & 89.0 & \checkmark \\
                             &    & 2 & 99.0 & \checkmark \\
        \midrule
        \multirow{4.5}{*}{15} & \multirow{2}{*}{10} & 1 & 9.9 & \checkmark \\
                             &    & 2 & 32.1 & \checkmark \\
        \cmidrule{2-5}
                             & \multirow{2}{*}{20} & 1 & 75.1 & \checkmark \\
                             &    & 2 & 96.7 & \checkmark \\
    \end{tabular}
    \caption{Percentage of trajectories leading to the death of the agent for the discriminative experiment, based on grid size (\(N\)), trajectory length (\(T\)) and number of enemies (\(E\)).}
    \label{tab:death_percentage}
\end{table}

\subsection{Results}
\label{subsec:experiments:results}

Results for the generative experiment are reported in Table~\ref{tab:generative_results} while the results for the discriminative task can be found in Table~\ref{tab:balanced_accuracy} and Table~\ref{tab:f1_score}. We report the mean and standard error for all the metrics.
We discuss the main findings of both experiments by highlighting the advantages of \mms.

\begin{table}[t]
    \setlength{\tabcolsep}{3pt}
    \centering
    \begin{tabular}{ccccc}
        \multicolumn{2}{c}{} & \multicolumn{3}{c}{\textbf{Methods}} \\
        \cmidrule{3-5}
        \textbf{Metric} & $N$ & VT & \dhmm & \mm \\
        \midrule
        \multirow{2}{*}{RE\ $(\downarrow)$} & 5  & \phantom{0}3.30 $\pm$ 0.04   & \phantom{0}4.97 $\pm$ 0.37   & \phantom{0}\textbf{3.32 $\pm$ 1.80} \\
                                           & 10 & \phantom{0}\textbf{2.23 $\pm$ 0.01}   & \phantom{0}4.66 $\pm$ 0.08   & \phantom{0}3.78 $\pm$ 0.07 \\
        \midrule
        \multirow{2}{*}{RA\ $(\uparrow)$}  & 5  & 91.39 $\pm$ 0.54 & 44.55 $\pm$ 5.75 & \textbf{97.17 $\pm$ 1.00}  \\
                                           & 10 & 30.62 $\pm$ 1.24 & \phantom{0}1.54 $\pm$ 0.00 & \textbf{89.63 $\pm$ 3.59}  \\
    \end{tabular}
    \caption{Reconstruction error (RE) and accuracy (RA) for the generative experiment on different grid sizes ($N{\times}N$). RE is multiplied by $1000$, and RA is in percentage.}
    \label{tab:generative_results}
\end{table}

\paragraph{Better generative consistency.}
Integrating knowledge about the environment is clearly advantageous to the generation process in terms of logical consistency, as can be seen from the reconstruction accuracies.
\mms significantly outperform both baselines in this regard, especially on larger grids.
In terms of reconstruction error, the variational transformer performs on-par or even better than \mms.
While this shows how transformers are very capable of optimising their losses, the sub-par reconstruction accuracy questions the degree of transfer from optimised solutions to desired solutions.
The \dhmm generally underperforms compared to both the VT and the \mm, illustrating the challenge of tackling sequential tasks without logic (\mm) or a longer time dependency (VT).

\begin{figure*}[t]
    \centering
    \includegraphics[width=\linewidth]{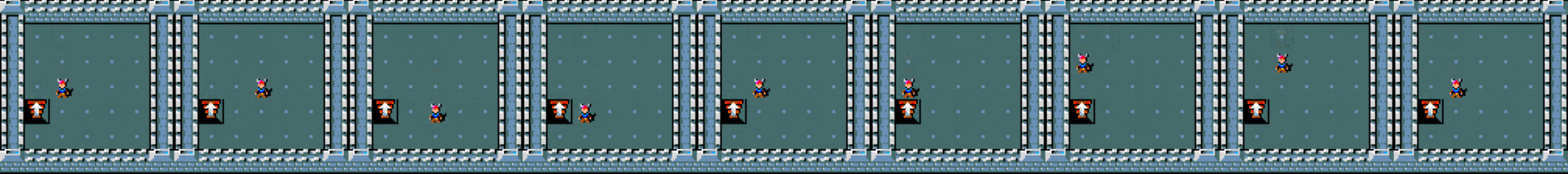}
    \caption{Generated trajectory for actions: right, down, left, up, left, up, right, down.}
    \label{fig:mario_actions}
\end{figure*}

\begin{figure*}[t]
    \centering
    \includegraphics[width=\linewidth]{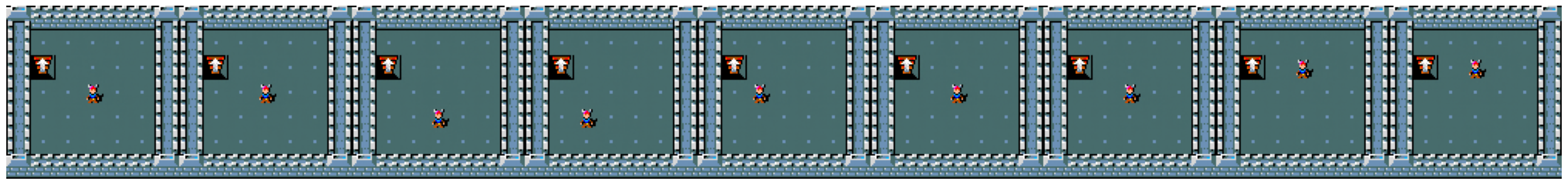}
    \caption{Generated trajectory for actions: \emph{right}, down, left, up, right, \emph{right}, up, \emph{right}; but with the test-time constraint that the area to the right of the start position should not be entered. When the agent is asked to move in the unsafe area (\ie actions in italics) it, instead, stays in the safe zone, and then it continues following the rest of the instructions.}
    \label{fig:mario_test}
\end{figure*}

\paragraph{Logical interpretability and intervenability.}
One of the biggest advantages of neurosymbolic generation is its ability to induce interpretable and intervenable logical consistency into subsymbolic generation. 
As an example, consider the generation in Figure~\ref{fig:mario_actions} where the generative model was asked to generate a trajectory for the agent, following a given sequence of actions, while adhering to the movement rules of the game.
Because the symbolic rules of the game are an inherent part of the generative model, \mms generate sequences that perfectly adhere to the mechanics of the game and the provided actions.
Other methods lack the necessary semantics or symbolic knowledge to fully guarantee this sort of logical consistency (Appendix~\ref{app:reproducibility}).
Moreover, \mms allow imposing constraints at test time in addition to the ones used during learning, which corresponds to zero-shot adherence to new queries (Figure~\ref{fig:mario_test}).

\paragraph{Scaling NeSy to non-trivial time horizons.}
NeSy methods are known for their scalability issues. 
When sequential generative settings are considered, the situation is even more dramatic. 
VAEL fails to perform inference on a single sequence of length $10$ even on a smaller grid of size $3{\times}3$, with $6h$ timeout.
On the contrary, we manage to perform inference and generative learning that does not deteriorate over time, even compared to the neural baselines. In fact, \mms perform a forward and backward pass over a batch of sequences in $\approx 0.25s$, for both grid sizes (more details in Appendix~\ref{app:reproducibility}). 
In the discriminative setting, the presence of a neural network as transition prevented us from applying any existing NeSy system, as they do not provide effective strategies to integrate neural networks except as perception.

\paragraph{Better out-of-distribution generalisation.}
Pivoting the attention to the discriminative experiment, we can see that \mms exploit their relational expressivity to perform well in all the out-of-distribution settings. 
Both the accuracy (Table~\ref{tab:balanced_accuracy}) and F1-score (Table~\ref{tab:f1_score}) paint a similar picture: the transformer is able to achieve better performance when staying in distribution ($N = 10$, $H = 10$ and $E = 1$). 
However, the OOD settings deteriorate the transformer's performance.
Only when the balance between positive and negative classes is closest to the balance of the training data, \ie when only increasing $N$ from 10 to 15 (Table~\ref{tab:death_percentage}), is the transformer able to keep a good accuracy.
In contrast, \mms show that their relational representations are much more robust to distribution shifts.
\dhmms land somewhere in the middle between transformers and \mms as their performance is always lower than \mms, but depending on the case they can be more robust than the transformers. Finally, notice that \dhmms cannot be applied to larger grid sizes, limiting their OOD capabilities.

\begin{table}[t]
    \centering
    \begin{tabular}{@{}p{0.5cm}@{} p{0.5cm}@{} p{0.5cm}@{} ccc@{}}
        \multicolumn{3}{c}{} & \multicolumn{3}{c}{\textbf{Methods}} \\
        \cmidrule{4-6}
        $N$ & $T$ & $E$ & Transformer & \dhmm & \mm \\
        \midrule
        \multirow{4.5}{*}{10} & \multirow{2}{*}{10} & 1 & \textbf{75.72 $\pm$ 1.35} & 59.88 $\pm$ 0.28 & 64.45 $\pm$ 1.10 \\
                            &                     & 2 & 68.61 $\pm$ 0.78 & 38.43 $\pm$ 0.21 & \textbf{78.13 $\pm$ 0.67} \\
                            \cmidrule{2-6}
                            & \multirow{2}{*}{20} & 1 & 50.40 $\pm$ 0.19 & 49.99 $\pm$ 0.39 & \textbf{67.66 $\pm$ 0.92} \\
                            &                     & 2 & 16.09 $\pm$ 1.33 & 49.33 $\pm$ 0.17 & \textbf{57.47 $\pm$ 0.56} \\
        \midrule
        \multirow{4.5}{*}{15} & \multirow{2}{*}{10} & 1 & \textbf{78.53 $\pm$ 3.09} & --- & 57.22 $\pm$ 1.78 \\
                            &                     & 2 & 67.85 $\pm$ 1.79 & --- & \textbf{75.57 $\pm$ 0.42} \\
                            \cmidrule{2-6}
                            & \multirow{2}{*}{20} & 1 & 50.47 $\pm$ 0.18 & --- & \textbf{71.85 $\pm$ 0.58}  \\
                            &                     & 2 & 41.54 $\pm$ 2.39 & --- & \textbf{77.13 $\pm$ 2.14} \\
    \end{tabular}
    \caption{Balanced accuracy (\%) for the discriminative experiment for grid sizes $N{\times}N$, with trajectory length $T$ and $E$ enemies. The first line is in-distribution performance, the rest is out of distribution. The \dhmm cannot be applied to bigger grid sizes.}
    \label{tab:balanced_accuracy}
\end{table}

\begin{table}[t]
    \centering
    \begin{tabular}{@{}p{0.5cm}@{} p{0.5cm}@{} p{0.5cm}@{} ccc@{}}
        \multicolumn{3}{c}{} & \multicolumn{3}{c}{\textbf{Methods}} \\
        \cmidrule{4-6}
        $N$ & $T$ & $E$ & Transformer & \dhmm & \mm \\
        \midrule
        \multirow{4.5}{*}{10} & \multirow{2}{*}{10} & 1 & \textbf{0.61 $\pm$ 0.02} & 0.25 $\pm$ 0.01 & 0.41 $\pm$ 0.03 \\
                            &                     & 2 & 0.59 $\pm$ 0.02 & 0.56 $\pm$ 0.00 & \textbf{0.79 $\pm$ 0.01} \\
                            \cmidrule{2-6}
                            & \multirow{2}{*}{20} & 1 & 0.02 $\pm$ 0.01 & 0.79 $\pm$ 0.01 & \textbf{0.92 $\pm$ 0.01} \\
                            &                     & 2 & 0.02 $\pm$ 0.01 & 0.98 $\pm$ 0.00 & \textbf{0.99 $\pm$ 0.00} \\
        \midrule
        \multirow{4.5}{*}{15} & \multirow{2}{*}{10} & 1 & \textbf{0.57 $\pm$ 0.03} & --- & 0.15 $\pm$ 0.02 \\
                            &                     & 2 & 0.52 $\pm$ 0.03 & --- & \textbf{0.65 $\pm$ 0.01} \\
                            \cmidrule{2-6}
                            & \multirow{2}{*}{20} & 1 & 0.02 $\pm$ 0.01 & --- & \textbf{0.78 $\pm$ 0.02} \\
                            &                     & 2 & 0.02 $\pm$ 0.01 & --- & \textbf{0.98 $\pm$ 0.01} \\
    \end{tabular}
    \caption{F1-Score for the discriminative experiment. This follows the same notation as Table~\ref{tab:balanced_accuracy}.}
    \label{tab:f1_score}
\end{table}

\section{Conclusion}
We introduced relational neurosymbolic Markov models (\mms), a powerful new class of relational probabilistic models that can incorporate neural networks beyond just perception modules.
These models are provided with a novel scalable and differentiable particle filtering technique for inference and learning, facilitating the bidirectional flow of information necessary for a proper neurosymbolic model~\ref{prop:nesy}.
Our empirical results show that the integration of relational symbolic knowledge into deep Markov models leads to significant improvements in generative and discriminative tasks~\ref{prop:discgen}, while also providing guarantees that neural models alone cannot achieve.
Importantly, we stressed the relational aspect of \mms by showing that purely neural models and even deep probabilistic models struggle to learn representations that generalise to unseen data and settings~\ref{prop:logic}.
While such generalisation behaviour is inherent to many neurosymbolic approaches, our experiments showed that \mms scale to sequential settings beyond the reach of existing NeSy systems~\ref{prop:sequential}.

Future work will focus on further applying \mms to new settings, such as reinforcement learning and
applications where continuous random variables are used differently from image generation, \eg physical systems.

\section{Acknowledgments}
This research has also received funding from the KU Leuven Research Funds (C14/24/092, STG/22/021),
from the Flemish Government under the "Onderzoeksprogramma Artificiële Intelligentie (AI) Vlaanderen" programme,
from the Wallenberg AI, Autonomous Systems and Software Program (WASP) funded by the Knut and Alice Wallenberg Foundation,
and from the European Research Council (ERC) under the European Union's Horizon Europe research and innovation programme (grant agreement n°101142702).

\bibliography{references}

\clearpage
\appendix
\section{Probabilistic Logic Programming}
\label{app:plp}

Logic programming has three main building blocks, being \emph{terms}, \emph{atoms} and \emph{rules}.
A term is the logic construct used to place the values one would like to reason over in a logical formula, like a constant \probloginline{c} or a variable \probloginline{V} in the simplest case.
Additionally, a term can also be formed recursively by applying a functor \probloginline{f} to a tuple of terms, i.e., something of the form \probloginline{f(t@$_1$@,...,t@$_K$@)}.
Next, atoms are relations that can be either true or false depending on their arguments and the given background knowledge. 
They are written using a predicate symbol \probloginline{q/@$K$@} of arity $K$ filled in with terms, e.g., \probloginline{q(t@$_1$@,...,t@$_K$@)}.
Atoms and terms are used in the definition of rules of the form \probloginline{h:- b@$_1$@,...,b@$_K$@}, where \probloginline{h} is an atom and each \probloginline{b@$_i$@} is a \emph{literal}, i.e. an atom or the negation of an atom.
We call \probloginline{h} the \emph{head} of the rule while the conjunction \probloginline{b@$_1$@,...,b@$_K$@} is the body of the rules.
If the body of the rule is logically true, then the rule expresses that the head of the rule is true as well by logical consequence.

\begin{example}[Logic Program]
The following logic program expresses the background knowledge necessary to find out if someone is a grandparent of someone else.
There is a shared knowledge base at the top in the form of two atoms that express that \probloginline{george} is the father of \probloginline{alice} and that \probloginline{alice} is the mother of \probloginline{william}.
The first two rules define the parent relation as being either the mother or the father of someone.
Finally, the last rule defines that \probloginline{X} is a grandparent of \probloginline{Y} if there exists an intermediate person \probloginline{Z} that is the parent of \probloginline{Y} and whose parent is \probloginline{X}.
\end{example}

{\fontsize{8.5}{9.5}\selectfont
\begin{problog}
father(george, alice). mother(alice, william).

parent(X, Y) :- father(X, Y).
parent(X, Y) :- mother(X, Y).

grandparent(X, Y) :- 
    parent(X, Z), parent(Z, Y).
\end{problog}
}

While atoms are either true or false in traditional logic programming, they can be probabilistically true or false when going to probabilistic logic programming.
That is, an atom can now be annotated with the probability that it is true.
For instance, \probloginline{0.42 :: father(george, alice)} expresses that one is only $42$ percent sure that \probloginline{george} is in fact the father of \probloginline{alice}.

Modern implementations of probabilistic logic programming allow for the probabilities of atoms to be parameterised by neural networks to achieve a neurosymbolic integration.
Apart from parametrising atoms representing finite random variables, these implementations have also been extended to the infinite domain~\citep{de2023neural}.
To facilitate the definition and inclusion of such atoms, two additional building blocks were added.
First, there is the \emph{neural distributional fact} (NDF), an expression of the form \probloginline{x ~ distribution(n@$_1$@,...,n@$_K$@)}.
Here, \probloginline{x} is a term representing a random variable distributed according to \probloginline{distribution} filled in with the numeric terms \probloginline{n@$_1$@,...,n@$_K$@}.
These numeric terms can be constant numerical values or the output of a neural network.
Second, to use neural distributional facts in a logical expression, which is always either true or false, \emph{probabilistic comparison formulae} (PCF) were also introduced. 
These are specific atoms that take the terms coming from a neural distributional fact as argument.

\begin{example}[DeepSeaProbLog program]
In the following piece of code, we show how NDFs and PCFs work together to write a probabilistic logic program that represents a simple model of the weather.
Two finite variables first model whether it is cloudy and what degree of humidity currently holds.
The \probloginline{temperature} NDF models a normal distribution with mean \probloginline{15} and standard deviation \probloginline{3}.
Because of the uncertain nature of these variables, the truth value of the atoms \probloginline{rain} and \probloginline{good_weather} will also be uncertain.
These atoms are defined under the NDFs; it is rainy when it is cloudy and humid, and the weather is good when it does not rain and the temperature is high or if it does rain, but the temperature is low.
{\fontsize{8.5}{9.5}\selectfont
\begin{problog}
cloudy ~ bernoulli(0.7)
humid ~ categorical([0.3, 0.5, 0.2], 
                    [dry, moist, wet]).
temperature ~ normal(15, 3).

rain :- cloudy, not (humid =:= dry).

good_weather :- not rain, temperature > 20.
good_weather :- rain, temperature < 0.
\end{problog}
}
\end{example}

Finally, let us provide some more details regarding the running example in the main body of the paper.

\begin{figure}
\centering
\begin{tikzpicture}
\node[sqnode, fill=celadon_blue, fill opacity=0.35] (image) {\texttt{Im}};
\node[circnode, fill=celadon_blue, fill opacity=0.35, below=2em of image] (player) {\texttt{P}};
\node[circnode, fill=celadon_blue, fill opacity=0.35, right=2em of image] (monster) {\texttt{M}};
\node[circnode, fill=celadon_blue, fill opacity=0.35, below right=0.2em and 0.8em of monster] (clumsy) {\texttt{C}};
\node[circnode, fill=celadon_blue, fill opacity=0.35, below=2em of monster] (hits) {\texttt{H}};
\node[circnode, fill=celadon_blue, fill opacity=0.35, below right=2em and 2.5em of hits] (game_over) {\texttt{G}};

\draw[thinpath] (image) -- node[midway, left] {$\varphi^\texttt{p}$} (player);
\draw[thinpath] (image) -- node[midway, above] {$\varphi^\texttt{m}$} (monster);
\draw[thinpath] (player) -- (hits);
\draw[thinpath] ([xshift=-0.25mm]monster.south) -- ([xshift=-0.25mm]hits);
\draw[thinpath] ([xshift=-0.25mm]clumsy.south) |- (hits);
\draw[thinpath] (hits) |- ([yshift=1mm]game_over.west);
\draw[thinpath] (player) |- ([yshift=-1mm]game_over.west);
\draw[thinpath] (monster.east) -| (game_over.north);
\end{tikzpicture}
\caption{Probabilistic graphical model view of the \dspl encoding in Figure~\ref{fig:nesy-game}.
}
\label{fig:nesy-game-pgm}
\end{figure}
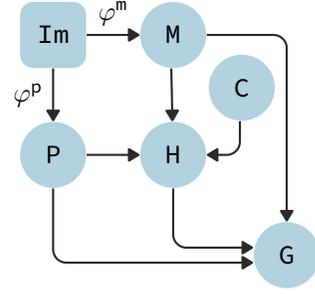

\begin{example}
Figure~\ref{fig:nesy-game-pgm} shows the probabilistic graphical model view of Example~\ref{ex:nesy-game}.
Notice how the two rules \texttt{H} (player hit by the monster) and \texttt{G} (game lost) are represented as binary random variables in the graphical model.
The logical relations between them and the other variables are implicitly defined by the topology of the model itself.
Specifically, the rules are encoded in the conditional probability tables, which we omit for brevity.
In mathematical terms, Equation~\ref{eq:nesy-inference} tells us that the probability that \probloginline{image.png} represents a lost game is
\begin{align*}
    &p(\mathproblog{game_over(image.png)}) = \\
    &\int\limits_{\mathrlap{(\texttt{P}, \texttt{M}, \texttt{C}) \models_{\texttt{H}} \texttt{g}}}
    \  p(\texttt{P} \mid \mathproblog{image.png}) p(\texttt{M} \mid \mathproblog{image.png}) p(\texttt{C})
    \ \mathrm{d}\texttt{P} \mathrm{d}\texttt{M} \mathrm{d}\texttt{C}, \notag
\end{align*}
with a small abuse of notation for the discrete variable \texttt{C}.
\end{example}

\section{Neurosymbolic Particle Filter}
\label{app:nesy-pf}
Let us focus on the global NeSy state variable $\mathbf{X}=(\mathbf{N},\mathbf{S})$ and how it evolves over time. 
The recursive equation of a particle filter applied to a NeSy-MM $(\mathbf{X}, \mathbf{Z})$ computing the probability of a state $\mathbf{X}_{t+1}$ at time step $t+1$ given observations $\mathbf{Z}_{0:t + 1}$ from time steps $0$ to $t+1$ is
\begin{align*}
    &p_{\nn}(\mathbf{X}_{t + 1} \mid \mathbf{Z}_{0:t + 1})
    = \\
    &\frac{p_{\nn}(\mathbf{Z}_{t + 1}  \mid \mathbf{X}_{t + 1})
    }{
    p_{\nn}(\mathbf{Z}_{t + 1} \mid \mathbf{Z}_{0:t})
    }
    \int
    p_{\nn}(\mathbf{X}_{t + 1} \mid \mathbf{x}_{t}) p_{\nn}(\mathbf{x}_{t} \mid \mathbf{Z}_{0:t})
    \ \mathrm{d}\mathbf{x}_{t}. \notag
\end{align*}
Practical implementations of these recursive equations require three steps.
First, a set $\{\mathbf{x}_{t}^{(n)}\}_{n = 1}^{N}$ of $N$ samples is drawn from the current time $t$ distribution, i.e., $\mathbf{x}_{t}^{(n)} \sim  p_{\nn}(\mathbf{X}_{t} \mid \mathbf{Z}_{0:t})$. 
Then, this set is transitioned to the next time step via the transition distribution $p_{\nn}(\mathbf{X}_{t + 1} \mid \mathbf{x}_{t})$. Finally, each of these samples is reweighted according to the observation probabilities $p_{\nn}(\mathbf{Z}_{t + 1}  \mid \mathbf{X}_{t + 1})$.
The resulting set of weighted samples is therefore approximately distributed according to $p_{\nn}(\mathbf{X}_{t + 1} \mid \mathbf{Z}_{0:t + 1})$.

Instead of keeping a set of samples $\{\mathbf{x}_{t}^{(n)}\}_{n = 1}^{N}$ weighted by their observations $p_{\nn}(\mathbf{Z}_{0:t + 1} \mid \mathbf{x}_{t}^{(n)})$, practical particle filters use \emph{resampling} to avoid the sample set from collapsing to samples with very low probability. 
Concretely, they use the observation probabilities $p_{\nn}(\mathbf{Z}_{0:t + 1} \mid \mathbf{x}_{t}^{(n)})$ as the weights of a finite random variable with $N$ outcomes, one for each sample $\mathbf{x}_t^{(n)}$ and take $N$ samples from the distribution of this variable to use as the recursive set of samples for the next filtering step.
Unfortunately, while the original set of weights could be differentiated and used to approximate gradients for every sample, the introduction of this auxiliary finite random variable is not differentiable, preventing a resampling particle filter from directly being used in our setting.

\section{Gradient Estimation for \mms}
\label{app:gradients}

This section will detail the precise way in which we use discrete gradient estimation to provide unbiased approximate gradients for \mms.
For brevity and ease of exposition, we assume without loss of generality that each $\mathbf{X}_t$ is finite in nature.
Our exposition will be fully general by considering the problem of computing
\begin{align}\label{eq:general_grad_exp}
    \nabla_{\nn}
    \Exp{
        \mathbf{X}_{0:T} \sim p_{\nn}(\mathbf{X}_{0:T} \mid \mathbf{Z}_{0:T})
    }{
        f(\mathbf{X}_{0:T})
    }.
\end{align}
In particular, all cases described in Section~\ref{sec:nesy-mm} are covered by the expectation in this equation.
Consider for example the discriminative task of computing $p_{\nn}(y \mid \mathbf{n}_{0:T}, \mathbf{Z}_{0:T})$ where the target random variable $Y$ depends on the symbols $\mathbf{S}_t$ for every $0 \leq t \leq T$.
By definition of a probability, we can write
\begin{align}
    &p_{\nn}(y \mid \mathbf{n}_{0:T}, \mathbf{Z}_{0:T}) \\
    &=
    \Exp{
        \mathbf{S}_{0:T} \sim p_{\nn}(\mathbf{S}_{0:T} \mid \mathbf{n}_{0:T}, \mathbf{Z}_{0:T})
    }{
        \indicator_{\mathbf{S}_{0:T} \models y}
    }, \nonumber
\end{align}
and this expression is indeed an expectation like the one in Equation~\ref{eq:general_grad_exp}, modulo given values for the neural variables $\mathbf{N}_{0:T}$.

To provide an unbiased approximation of the gradient in Equation~\ref{eq:general_grad_exp}, we start by applying the Log-Derivative Trick~\citep{williams1992simple}, resulting in the sum of the two terms
\begin{align}\label{eq:grad_term_1}
    \Exp{
        \mathbf{X}_{0:T} \sim p_{\nn}(\mathbf{X}_{0:T} \mid \mathbf{Z}_{0:T})
    }{
        \nabla_{\nn} f(\mathbf{X}_{0:T})
    },
\end{align}
and
\begin{align}\label{eq:grad_term_2}
    \Exp{
        \mathbf{X}_{0:T} \sim p_{\nn}(\mathbf{X}_{0:T} \mid \mathbf{Z}_{0:T})
    }{
        f(\mathbf{X}_{0:T})
        \nabla_{\nn}
        \log 
        p_{\nn}(\mathbf{X}_{0:T} \mid \mathbf{Z}_{0:T})
    }.
\end{align}
In all that follows, we will ignore the first term (Equation~\ref{eq:grad_term_1}) as $\nabla_{\nn} f(\mathbf{x}_{0:T})$ can easily be computed for any instance $\mathbf{x}_{0:T}$ using automatic differentiation algorithms.
The second term is where many of the problems in gradient estimation arise.
While the infamously high variance of a direct Monte Carlo estimate of Equation~\ref{eq:grad_term_2} is problematic, we additionally have to deal with the fact that the gradient $\nabla_{\nn}\log p_{\nn}(\mathbf{X}_{0:T} \mid \mathbf{Z}_{0:T})$ could prove problematic.
To minimise the former problem of variance, we use the unbiased RLOO estimator~\citep{kool2019buy} of Equation~\ref{eq:grad_term_2}.
\begin{align}\label{eq:nesy_rloo}
    \frac{1}{N - 1} \sum_{i = 1}^N
    \left(
        f(\mathbf{x}_{0:T}^{(i)})
        -
        \overline{f}
    \right)
    \nabla_{\nn}
    \log
    p_{\nn}(\mathbf{x}_{0:T}^{(i)} \mid \mathbf{Z}_{0:T}),
\end{align}
where $\overline{f} = \frac{1}{N} \sum_{j = 1}^N f(\mathbf{x}_{0:T}^{(j)})$ is an empirical estimate of the average of $f$ with respect to $p_{\nn}(\mathbf{X}_{0:T} \mid \mathbf{Z}_{0:T})$.

Next, we can start decomposing $\nabla_{\nn}\log p_{\nn}(\mathbf{x}_{0:T}^{(i)} \mid \mathbf{Z}_{0:T})$ and show that we are able to compute it exactly.
Notice how the joint distribution follows a recursive factorisation similar to the recursive integration of the RBPF (Equation~\ref{eq:rbpf_recursion})
\begin{align}
    &p_{\nn}(\mathbf{x}_{0:T}^{(i)} \mid \mathbf{Z}_{0:T}) \\
    &=
    p_{\nn}(\mathbf{x}_{T}^{(i)} \mid \mathbf{x}_{T - 1}^{(i)}, \mathbf{Z}_{t + 1})
    p_{\nn}(\mathbf{x}_{0:T - 1}^{(i)} \mid \mathbf{Z}_{0:T - 1}). \nonumber
\end{align}
Because we assume that $p_{\nn}(\mathbf{X}_{T} \mid \mathbf{x}_{T - 1}^{(i)}, \mathbf{Z}_{t + 1})$ can be computed exactly, as necessitated by the use of our RBPF, it follows that we can also exactly compute the gradient of $p_{\nn}(\mathbf{x}_{T}^{(i)} \mid \mathbf{x}_{T - 1}^{(i)}, \mathbf{Z}_{t + 1})$.
Consequently, we can simply apply automatic differentiation on 
\begin{align}
    &\log p_{\nn}(\mathbf{x}_{0:T}^{(i)} \mid \mathbf{Z}_{0:T}) \\
    &=
    \sum_{t = 0}^T 
    \log p_{\nn}(\mathbf{x}_{t}^{(i)} \mid \mathbf{x}_{t - 1}^{(i)}, \mathbf{Z}_{t + 1}), \nonumber
\end{align}
to compute the gradient $\nabla_{\nn}\log p_{\nn}(\mathbf{x}_{0:T}^{(i)} \mid \mathbf{Z}_{0:T})$.
Hence, as alluded to in Section~\ref{sec:method}, we can now clearly see how the assumption of our RBPF makes it possible to apply gradient estimation following Equation~\ref{eq:nesy_rloo}.

To end this section, we further elaborate on the example given in Equation~\ref{eq:log_trick_npf} as is an interesting special case of Equation~\ref{eq:nesy_rloo}.
Indeed, gradients of the marginal $p_{\nn}(\mathbf{X}_{t} \mid \mathbf{Z}_{0:t})$ often appear when the target query variable $Y$ only depends on a single time step.
For example, assume we are interested in knowing what the probability is that the game from Example~\ref{ex:nesy-game} is over \emph{now} given that we have observed a series of hits in the \emph{past}.
In this case, we would have an expectation of the form
\begin{align}
    \Exp{
        \mathbf{X}_{T} \sim p_{\nn}(\mathbf{X}_{T} \mid \mathbf{Z}_{0:T})
    }{
        f(\mathbf{X}_{T})
    },
\end{align}
leading to an application of RLOO that requires $\nabla_{\nn} p_{\nn}(\mathbf{x}_{T}^{(i)} \mid \mathbf{Z}_{0:T})$.
Since we only have samples from the joint distribution $p_{\nn}(\mathbf{X}_{0:T} \mid \mathbf{Z}_{0:T})$, we approximate $\nabla_{\nn} p_{\nn}(\mathbf{x}_{T}^{(i)} \mid \mathbf{Z}_{0:T})$ via a Monte Carlo estimate of Equation~\ref{eq:log_trick_npf}.
To again mitigate the high variance of the recursive term $\Exp{
        \mathbf{X}_{T - 1} 
    }{
        p_{\nn}(\mathbf{x}_{T}^{(i)} \mid \mathbf{X}_{T - 1}, \mathbf{Z}_{T})
        \nabla_{\nn}
        \log p_{\nn}(\mathbf{X}_{T - 1} \mid \mathbf{Z}_{0:T})
    }$
we apply RLOO \emph{recursively} through the expression
\begin{align}
    \frac{1}{N - 1} \sum_{j = 1}^N
    &\Big(
        p_{\nn}(\mathbf{x}_{T}^{(i)} \mid \mathbf{x}_{T - 1}^{(j)}, \mathbf{Z}_{T})
        - \\
        &\overline{p}_{\nn}(\mathbf{x}_{T}^{(i)} \mid \mathbf{Z}_{T})
    \Big)
    \nabla_{\nn}
    \log p_{\nn}(\mathbf{x}_{T - 1}^{(j)} \mid \mathbf{Z}_{0:T - 1}), \nonumber
\end{align}
where now 
\begin{align}
    \overline{p}_{\nn}(\mathbf{x}_{T}^{(i)} \mid \mathbf{Z}_{T}) = \frac{1}{N} \sum_{k = 1}^N p_{\nn}(\mathbf{x}_{T}^{(i)} \mid \mathbf{x}_{T - 1}^{(k)}, \mathbf{Z}_{T}),
\end{align}
with $x_{T - 1}^{(k)} \sim p_{\nn}(\mathbf{X}_{T - 1} \mid \mathbf{Z}_{0:T - 1})$.

\section{Reproducibility}
\label{app:reproducibility}

\subsection{Architectures}

\paragraph{Generative Task.}
The variational transformer architectures consist of a separate initial convolutional VAE architecture coupled with a transformer decoder that autoregressively generated the next images in the sequence. This transformer has a causal self-attention layer with 8 heads and 64 keys, followed by a cross attention layer with the same parameters. After these layers, the decoder portion of the VAE is used to generate the images.
The NeSy-MM model used multiple smaller networks as it is naturally decomposed.
Concretely, there is a small convolutional neural network that classifies the initial location of the agent from the first image into either $(5 + 2) ^2 = 49$ or $(10 + 2)^2 = 144$ classes, depending on the grid size.
Additionally, a small convolutional encoder network encodes the same initial image into a two-dimensional Gaussian distribution.
The NeSy-MM moves the location of the agent according to the rules of MiniHack for a given set of actions, resulting in an estimated distribution for the agent at every subsequent time step.
A final convolutional decoder, the same as used by the transformer architecture, then generates an image from the encoding of the initial image together with the planned location of the agent for every time step.
The deep Markov model (Deep-MM) uses the exact same setup, only replacing the logical transitions by small neural networks with two hidden layers of size $64$ and $32$. 

\paragraph{Discriminative Task.}
Our transformer architecture follows the usual pattern of an encoder-free transformer. 
That is, it has a decoder component that operates on a sequence of embedding vectors of size 32.
At the start, there is only one embedding containing the two-dimensional starting location of the agent as the first two components, followed by a series of zeroes.
The decoder then autoregressively computes the next embedding from all previous embeddings by applying, in sequence, a dropout operation with probability 0.1, a causal self-attention layer with 8 heads and key dimension 64, and finally a cross-attention layer with as context the incoming action and observation on whether the agent was hit or not.
The cross-attention layer again has 8 heads and key dimension 64 and both attention layers also use dropout internally with a probability of 0.1.
To provide the final sequence classification, a simple MLP network with 2 hidden layers of sizes 64 and 32 with ReLU activations is used.
It takes the final embedding vector as input such that it can deal with sequences of varying length, \ie it is relational in time.
The dense output layer is of dimension 1 and uses a sigmoid activation, predicting the probability that the agent dies in the trajectory or not.

Our \mm model and \dhmm are again close in terms of architecture, but differ in their use of neural networks.
Both models take the initial location of the agent as input and estimate the location of all enemies by a uniform prior.
They transition these locations to future time steps using actions.
For the enemies, the actions are not given and this is where the \mm uses a simple MLP with two hidden ReLU layers of size 64 and 32 and an output log softmax layer of size 8, as the enemies can move vertically, horizontally, and diagonally.
In short, the \mm transitions the agent from its previous location using the given action, while the enemies are transitioned from their sampled previous locations and \emph{predicted} actions.
\dhmms use a neural network to immediately predict the next location for both the agent and the enemy.
In case of the agent, the neural net takes both location and given action as input.
The architecture is the same as the action network of the \mm, albeit with an output of size $(N + 2)^2$ where $N$ is the grid size, predicting a distribution over grid cells.
Here we can also see why the \mm can tackle larger grid sizes while the \dhmm can not.
Our \mm uses a relational logic transition to move entities in the world (Appendix~\ref{app:subsec:rules}) while the \dhmm instead uses a neural network that necessarily has a fixed output dimension in terms of the grid size.
Next, both models \emph{exactly} condition, in the probabilistic sense, the predicted distribution of future locations on the incoming evidence whether the agent was hit or not.
For the \mm, the observation function is again logical, meaning it computes the probability that the agent is hit, given the agent location and all enemy locations.
This involves knowing the probability that an attack from an enemy succeeds.
Since this information can be seen as part of the behaviour of the monster, we do not give it as input to our model and instead replace it by a learnable parameter.
For the \dhmm, the observation function is completely replaced by a neural network with two hidden ReLU layers of size 64 and 32 followed by a dense log softmax output layer modelling the log probability. 
That is, computing the probability that an enemy hits the agent from their locations is computed by a neural net for each enemy separately and then combined into the total probability of hitting following the correct expression for the probability of the union of multiple events.
Finally, both models explicitly model the health of the agent and subtract an estimate of the damage based on the computed probability of hitting.
The final probability that the agent is dead predicted by the models is then given by the frequency of sampled trajectories where the agent dies.

\subsection{Rules of NetHack}
\label{app:subsec:rules}
Our \mm relies on relational logical knowledge. In this section, we describe in detail the knowledge that we included in our model.

\paragraph{Generative Task.} The knowledge necessary in this case is very simple and can be summarised by the following logic program.

{\fontsize{8.5}{9.5}\selectfont
\begin{problog}
agent(X, Y, T) ~ detector(Image,T).

action(A, T) ~ categorical(
    [0.25, 0.25, 0.25, 0.25],
    [up, down, left, right]
).

agent(X,Y+1,T) :- 
    action(up,T-1), agent(X,Y,T-1).
agent(X,Y-1,T) :- 
    action(down,T-1), agent(X,Y,T-1).
agent(X+1,Y,T) :- 
    action(right,T-1), agent(X,Y,T-1).
agent(X-1,Y,T) :- 
    action(left,T-1), agent(X,Y,T-1).
\end{problog}
}

Where, on the first line, we say that the agent location at time \probloginline{T} is given by a neural detector. Then, we just describe that there are four possible mutually exclusive actions (\probloginline{up}, \probloginline{down}, \probloginline{left}, and \probloginline{right}). Finally, we describe the effect of each action on the agent's location.

\paragraph{Discriminative Task.}
In this experiment, we do not get the agent's location by applying a neural detector to an image.
Instead, the exact and deterministic initial symbolic location is given.
The remainder of the logic for how the agent transitions is the same as in the generative task

For the enemies, the setup is analogous, at least in terms of how they transition.
The difference being that the enemies' location is initially completely uncertain, modelled by a uniform categorical over the entire grid.
Additionally, the actions are not uniformly sampled, but are predicted by the neural network that we are trying to optimise.

To eventually deduce whether the agent has died or not, we need to keep track of the agent's health and model the damage the agent takes
{\fontsize{8.5}{9.5}\selectfont
\begin{problog}
damage(T, Damage) ~ categorical(
    [0.25, 0.25, 0.25, 0.25],
    [1, 2, 3, 4]
)

agent_hp(0, 12).
agent_hp(T, HP) :-
    agent_hp(T-1, HP),
    not hit(T).
agent_hp(T, HP - Damage) :-
    agent_hp(T-1, HP),
    damage(T, Damage),
    hit(T).
\end{problog}
}

At the beginning when \probloginline{T} is $0$, the agent has $12$ hitpoints (HP) that decreases by the amount \probloginline{Damage} if there is a \probloginline{hit}.
The damage value is dependent on the enemy type.
In our case, we used the \nethack \texttt{imp} that has a claw attack with 1d4 damage.
Hence, the damage is modelled by a uniform categorical variable over the domain \probloginline{[1, 2, 3, 4]}.
Furthermore, a \probloginline{hit} can occur only if the enemy is in one of the $8$ cells around the agent, because the claw attack is a melee attack:
{\fontsize{8.5}{9.5}\selectfont
\begin{problog}
hit(T) ~ bernoulli(t(_)) :-
    agent(Xa, Ya, T), A = [Xa, Ya],
    enemy(Xe, Ye, T), E = [Xe, Ye],
    distance(A, E, D), D = 1.
\end{problog}
}

Notice that the \probloginline{hit} variable is Bernoulli distributed and the parameters are learned as indicated by the predicate \probloginline{t(_)}.
This predicate represents a learnable variable.
In this case the probability that a hit succeeds, which we do not assume to know as we consider it as part of the unknown behaviour of the enemy.
Therefore, we do not know when the hit is successful, but we observe the \probloginline{hit} variable during training.
Finally, we need a rule to understand when the agent is dead:
{\fontsize{8.5}{9.5}\selectfont
\begin{problog}
agent_dead(T) :-
    agent_hp(T, HP),
    HP =< 0.
\end{problog}
}

\subsection{Setup and hyperparameters}
\label{app:hyperparameters}
We ran the experiments on an NVIDIA P100 SXM2@1.3 GHz (16 GB HBM2) GPU, coupled with an Intel Xeon Gold 6140 CPU@2.3 GHz (Skylake) and 192 GB of RAM.
From the software perspective, the machine we used runs the Rocky Linux 8.9 (Green Obsidian) operating system. Moreover, we ran all the experiments in a Python 3.10.4 environment with the packages listed in the \texttt{requirements.txt} file available in the codebase. We report in Table~\ref{tab:python-packages} the exact versions used to produce the results in the paper.
All experiments were repeated 5 times on this setup for our method and all baselines. Results are reported using averages and standard errors. For the generative experiments on grid size $10$, we downscale the images by a factor of 2 to use 8 pixels per cell instead of the standard 16 pixels per cell.
This downscaling is due to the memory limitation of the GPU we had available and was applied for all the models.
The hyperparameters were obtained via a separate grid search using a held-out validation set. Table~\ref{tab:hyperparameters-generative} and Table~\ref{tab:hyperparameters-discriminative} report the tested values, together with the optimal ones used to produce the results in the paper. 
Adam~\citep{kingma2015adam} was used as the optimiser for all methods. 
We used seeds to remove as much randomness as possible from the training process. The five runs for the generative experiment were obtained all with seed $42$, while the ones for the discriminative experiment have been obtained with seeds from $0$ to $4$. The reason for this difference is merely technical: in the discriminative case, the evaluation is done in a second phase because of the several out-of-distribution settings, so we needed to easily distinguish the 5 runs and the seed number was the easiest choice.

The datasets used in the paper are generated with the \texttt{generator.py} scripts present in the \texttt{data/} folder of each experiment. The seed used for the generation is $0$. All the other parameters are described in Section~\ref{sec:experiments}. Running the training scripts with the default parameters automatically creates the datasets used to produce the results in the paper. With this information, the generation of the datasets is fully deterministic and perfectly reproducible. Train, test, and validation sets contain, respectively $5000$, $1000$, and $500$ trajectories.

\begin{table}[t]
\centering
\begin{tabular}{ll}
\textbf{Package}              & \textbf{Version} \\ 
\hline\rule{0pt}{2ex}%
tensorflow[and-cuda]          & \texttt{2.13.1} \\ 
tensorflow\-probability       & \texttt{0.19.0}   \\ 
gym                           & \texttt{0.23.0} \\ 
minihack                      & \texttt{0.1.6}    \\ 
nle                           & \texttt{0.9.0}    \\ 
einops                        & \texttt{0.8.0}    \\ 
wandb                         & \texttt{0.13.5}   \\ 
matplotlib                    & \texttt{3.8.0}    \\ 
\end{tabular}
\caption{Required packages and their versions.}
\label{tab:python-packages}
\end{table}

\begin{table}[t]
\centering
\begin{tabular}{cccc}
& \multicolumn{3}{c}{\textbf{Methods}} \\
\cmidrule{2-4}
$N$ & VT & \dhmm & \mm \\
\hline\rule{0pt}{2ex}%
5                  & 385.48 $\pm$ 2.79 & 1392.48 $\pm$ 6.83 & 726.17 $\pm$ 3.40 \\
10                 & 409.32 $\pm$ 1.97  & \phantom{1}410.15 $\pm$ 0.31  & 399.61 $\pm$ 4.83 \\
\end{tabular}
\caption{Total training time (s) for the generative experiment, for grid sizes $N{\times}N$ and sequence length $10$.}
\label{tab:training_times_generative}
\end{table}

\begin{table*}[t]
\centering
\begin{tabular}{lllll}
& & \multicolumn{3}{c}{\textbf{Optimal Values}} \\
\cmidrule{3-5}
\textbf{Hyperparameter}      & \textbf{Tested Values} & Transformer          & \dhmm             & \mm              \\ 
\hline\rule{0pt}{2ex}%
\texttt{n\_samples}          & 5, 10, 20              & -                    & 10                & 5                \\
\texttt{latent\_dim}         & 2, 4, 8                & -                    & 2                 & 2                \\
\texttt{dropout}             & 0.1, 0.2, 0.4, 0.5     & 0.1                  & 0.2               & 0.2              \\
\texttt{batch\_size}         & -                      & 10                   & 10                & 10               \\
\texttt{n\_epochs}           & 15, 30, 50, 100, 200   & 100                  & 30                & 30               \\
\texttt{beta}                & 1, 10, 50, 100, 150, 200, 300   & 1                    & 50                & 300              \\
\texttt{learning\_rate}      & 0.1, 0.03, 0.01, 0.001, 1e-4 & 1e-4                 & 0.001             & 0.001            \\           \\
\end{tabular}
\caption{Hyperparameters for the generative experiment. Tested values, and their optimal values for Transformer, \dhmm, and \mm. For the experiment with grid size 10, we changed: for \mm and \dhmm, \texttt{n\_samples} to 20, \texttt{batch\_size} to 5, and \texttt{n\_epochs} to 15; for the transformer \texttt{beta} to 50.}
\label{tab:hyperparameters-generative}
\end{table*}

\begin{table*}[t]
\centering
\begin{tabular}{lllll}
& & \multicolumn{3}{c}{\textbf{Optimal Values}} \\
\cmidrule{3-5}
\textbf{Hyperparameter}      & \textbf{Tested Values} & Transformer          & \dhmm             & \mm              \\ 
\hline\rule{0pt}{2ex}%
\texttt{n\_samples}          & 100, 1000              & -                    & 100               & 1000             \\
\texttt{dropout}             & 0.1, 0.2, 0.4          & 0.1                  & -                 & -                \\
\texttt{batch\_size}         & -                      & 50                   & 10                & 50               \\
\texttt{n\_epochs}           & 15, 30, 50, 100, 200   & 50                   & 20                & 100              \\
\texttt{learning\_rate}      & 0.01, 0.001, 1e-4      & 1e-3                 & 1e-3              & 1e-3             \\
\end{tabular}
\caption{Hyperparameters for the discriminative experiment. Tested values, and their optimal values for Transformer, \dhmm, and \mm.}
\label{tab:hyperparameters-discriminative}
\end{table*}

\subsection{Generation}
\begin{figure*}[t]
    \centering
    \includegraphics[width=\linewidth]{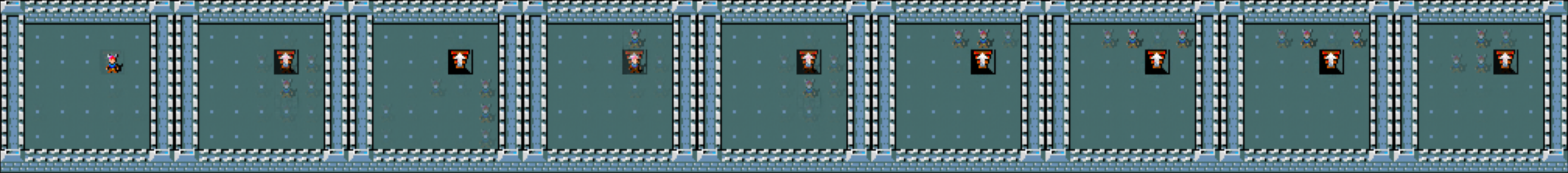}

    \vspace{2mm}
    \includegraphics[width=\linewidth]{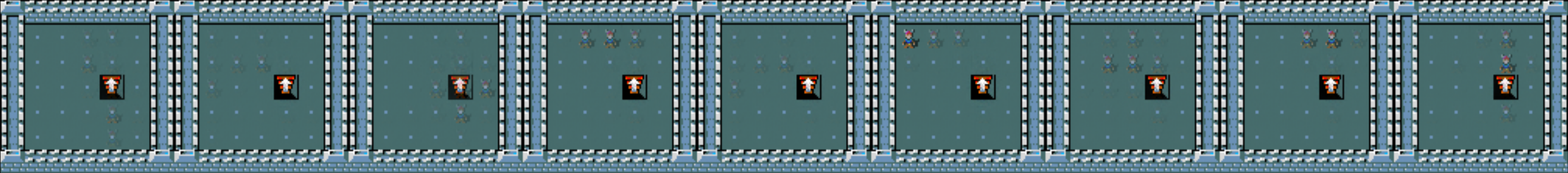}

    \vspace{2mm}
    \includegraphics[width=\linewidth]{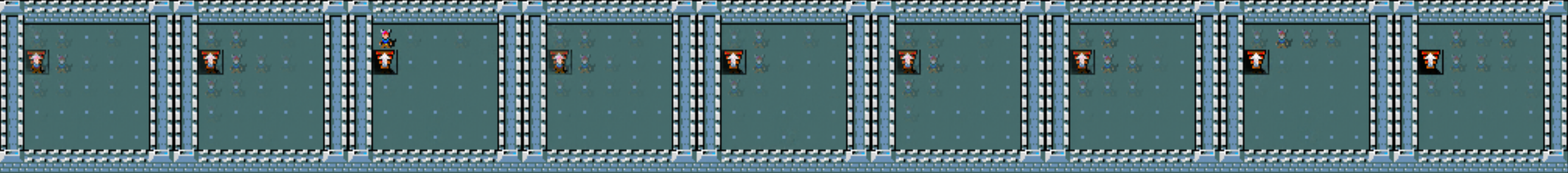}
    \caption{Generated trajectory for actions: right, down, left, up, right, up, left, down using the \dhmm.}
    \label{fig:deephmm_showcase}
\end{figure*}

\begin{figure*}[t]
    \centering
    \includegraphics[width=\linewidth]{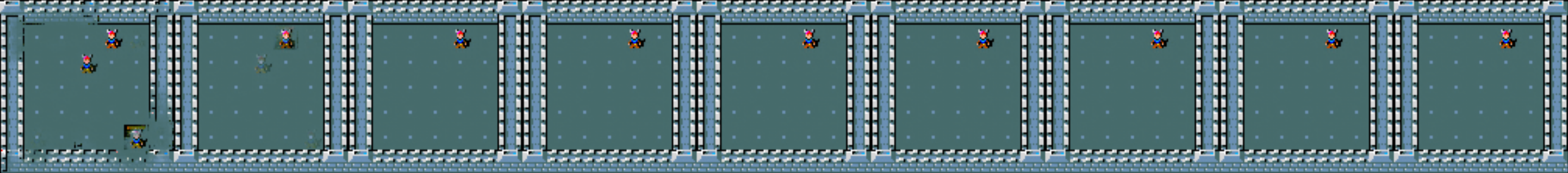}
    
    \vspace{2mm}
    \includegraphics[width=\linewidth]{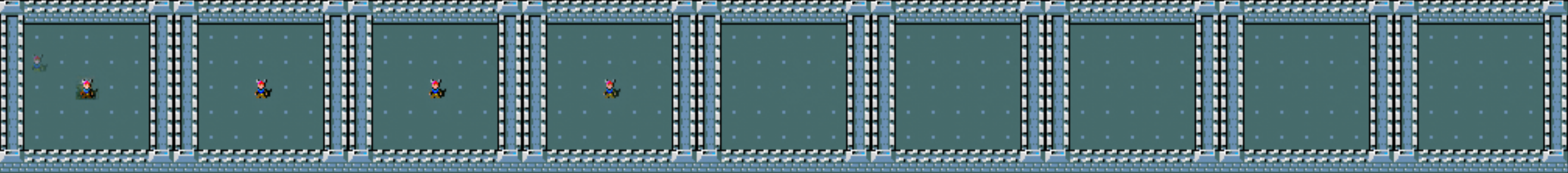}
    
    \vspace{2mm}
    \includegraphics[width=\linewidth]{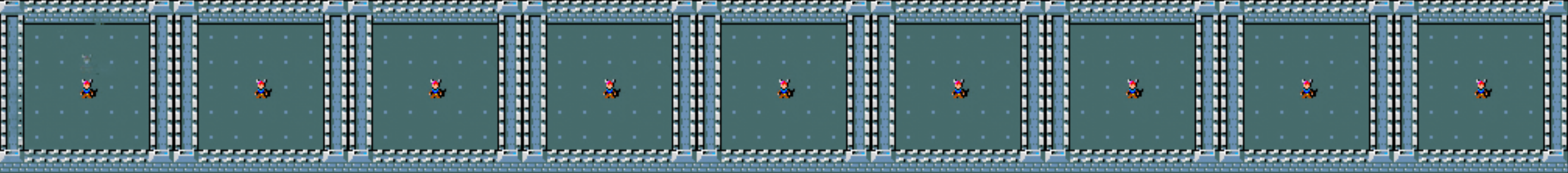}
    \caption{Generated trajectory for actions: right, down, left, up, right, up, left, down using the variational transformer.}
    \label{fig:vt_showcase}
\end{figure*}

In this section, we showcase the generative quality of \dhmms (Figure~\ref{fig:deephmm_showcase}) and variational transformers (Figure~\ref{fig:vt_showcase}), compared to the \mm generation from Figure~\ref{fig:mario_actions}.
Both the transformer and \dhmm might be able to provide reasonable test metrics, but they prove to be rather poor generative models.
The \dhmm provides clean generations of the background, but it fails to capture the exact location of the agent and often generates the agent in a superposition of different locations.
Moreover, the agent can jump around the world because the transitions are not guaranteed to follow the rules of the game.
The transformer also provides, most of the time, crisp generations of the background, but it also fails to properly learn the correct transition function and sometimes shows artefacts.
In many cases, even if the initial generation is clear, the transformer is unable to move the agent according to the required actions.
In contrast, \mms provide clean generation that follow the required actions and adhere to the rules of \nethack.
The reader can try to generate more images, to further confirm our findings, with the Jupyter notebook provided in the submitted code material.

\subsection{Training Time}
Table~\ref{tab:training_times_generative} shows the total training times for the generative experiment. For the discriminative experiment, where we train the model only on grid size $10{\times}10$, sequence length $10$, and $1$ enemy, the times are: $161.79 \pm 1.19$s for the transformer, $805.34 \pm 2.62$s for the \dhmm, and $254.53 \pm 9.11$s for our \mm.
Note how the \dhmm model takes considerably longer to train and delivers consistently lower performance.
For the transformer baselines, we always trained until convergence or until reaching the hardware's limitations.
Although these models generally train faster, they still underperform as demonstrated in the paper.
The slightly longer training times for our \mms can be attributed to the fact we perform sample-wise exact reasoning.

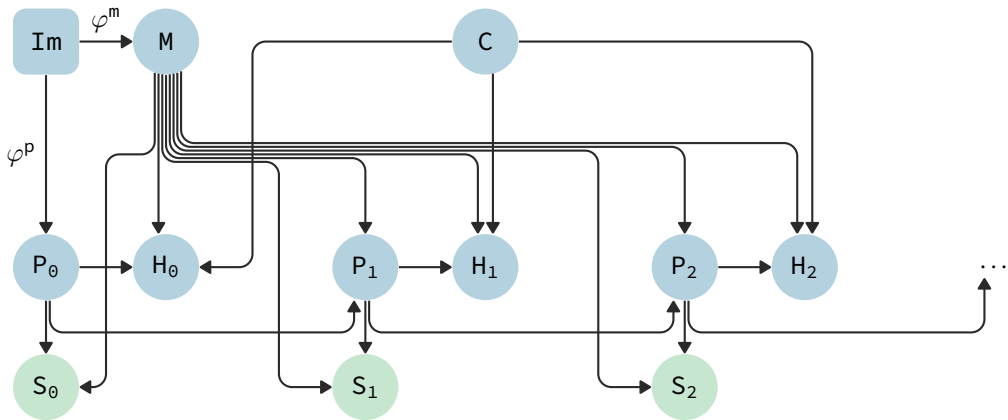
\begin{figure*}
\centering
\begin{tikzpicture}
\node[sqnode, fill=celadon_blue, fill opacity=0.35] (image) {\texttt{Im}};
\node[circnode, fill=celadon_blue, fill opacity=0.35, below=6em of image] (player0) {$\texttt{P}_\texttt{0}$};
\node[circnode, fill=celadon_blue, fill opacity=0.35, right=2em of image] (monster) {\texttt{M}};
\node[circnode, fill=celadon_blue, fill opacity=0.35, below=6em of monster] (hits0) {$\texttt{H}_\texttt{0}$};
\node[circnode, fill=grassy_green, fill opacity=0.35, below=2em of player0] (safe0) {$\texttt{S}_\texttt{0}$};

\node[circnode, fill=celadon_blue, fill opacity=0.35, right=5em of hits0] (player1) {$\texttt{P}_\texttt{1}$};
\node[circnode, fill=celadon_blue, fill opacity=0.35, right=2em of player1] (hits1) {$\texttt{H}_\texttt{1}$};
\node[circnode, fill=grassy_green, fill opacity=0.35, below=2em of player1] (safe1) {$\texttt{S}_\texttt{1}$};

\node[circnode, fill=celadon_blue, fill opacity=0.35, above=6em of hits1] (clumsy) {\texttt{C}};

\node[circnode, fill=celadon_blue, fill opacity=0.35, right=5em of hits1] (player2) {$\texttt{P}_\texttt{2}$};
\node[circnode, fill=celadon_blue, fill opacity=0.35, right=2em of player2] (hits2) {$\texttt{H}_\texttt{2}$};
\node[circnode, fill=grassy_green, fill opacity=0.35, below=2em of player2] (safe2) {$\texttt{S}_\texttt{2}$};

\node[right=5em of hits2] (dots) {\ldots};


\draw[thinpath] (image) -- node[midway, left] {$\varphi^\texttt{p}$} (player0);
\draw[thinpath] (image) -- node[midway, above] {$\varphi^\texttt{m}$} (monster);
\draw[thinpath] (player0) -- (hits0);
\draw[thinpath] (player0) -- (safe0);
\draw[thinpath] 
    ([xshift=0.5mm]player0.south) -- 
    ([xshift=0.5mm, yshift=-1.2em]player0.south) -|
    ([xshift=-1.5mm]player1.south);
\draw[thinpath] 
    ([xshift=-1.5mm, yshift=0.3mm]monster.south) --
    ([xshift=-1.5mm, yshift=-3em]monster.south) -|
    ([xshift=1em]safe0.east) --
    (safe0.east);
\draw[thinpath] ([xshift=-1mm, yshift=0.2mm]monster.south) -- ([xshift=-1mm]hits0.north);
\draw[thinpath] 
    ([xshift=0.5mm, yshift=0.1mm]monster.south) -- 
    ([xshift=0.5mm, yshift=-3em]monster.south) -|
    ([xshift=-1mm]hits1.north);
\draw[thinpath] 
    ([xshift=1.5mm, yshift=0.3mm]monster.south) -- 
    ([xshift=1.5mm, yshift=-3em + 1mm]monster.south) -|
    (player2.north);
\draw[thinpath] 
    ([xshift=0mm]monster.south) -- 
    ([xshift=0mm, yshift=-3em - 0.5mm]monster.south) -|
    (player1.north);
\draw[thinpath] (player1) -- (hits1);
\draw[thinpath] (player1.south) -- (safe1.north);
\draw[thinpath] 
    ([xshift=-0.5mm, yshift=0.1mm]monster.south) -- 
    ([xshift=-0.5mm, yshift=-3em - 1mm]monster.south) -|
    ([xshift=-2em]safe1.west) --
    (safe1.west);
\draw[thinpath] 
    ([xshift=0.5mm]player1.south) -- 
    ([xshift=0.5mm, yshift=-1.2em]player1.south) -|
    ([xshift=-1.5mm]player2.south);
    
\draw[thinpath] 
    ([xshift=2mm, yshift=0.5mm]monster.south) -- 
    ([xshift=2mm, yshift=-3em + 1.5mm]monster.south) -|
    ([xshift=-1mm]hits2.north);
\draw[thinpath] (player2) -- (hits2);
\draw[thinpath] (player2) -- (safe2);
\draw[thinpath] 
    ([xshift=1mm, yshift=0.2mm]monster.south) -- 
    ([xshift=1mm, yshift=-3em + 0.5mm]monster.south) -|
    ([xshift=-2em]safe2.west) --
    (safe2.west);
\draw[thinpath] 
    ([xshift=0.5mm]player2.south) -- 
    ([xshift=0.5mm, yshift=-1.2em]player2.south) -|
    ([xshift=-1.5mm]dots.south);
    
\draw[thinpath] ([xshift=1mm, yshift=0.2mm]clumsy.south) -- ([xshift=1mm]hits1.north);
\draw[thinpath] (clumsy.east) -| ([xshift=1mm]hits2.north);
\draw[thinpath] 
    (clumsy.west) -|
    ([xshift=2em]hits0.east) --
    (hits0.east);
\end{tikzpicture}
\caption{Rolled-out graphical model of Figure~\ref{fig:markov-game}, from Example~\ref{ex:markov-game}. We leave out the query node \texttt{G} because the edges to connect it would just cram the picture.}
\label{fig:markov-game:rolled-out}
\end{figure*}

\section{Reproducibility Checklist}
We answer affirmatively to the checklist point ``\emph{All novel datasets introduced in this paper are included in a data appendix}''. However, we do not include the actual datasets ($\approx 70$GB), but the code to generate them and the seed used. With this information, the dataset is deterministically generated and can be perfectly reproduced (see Appendix~\ref{app:hyperparameters}).

\end{document}